\documentclass{article}

\usepackage{microtype}
\usepackage{graphicx}
\usepackage{subfigure}
\usepackage{colortbl,hhline}
\usepackage{colortbl}
\usepackage{booktabs} 
\usepackage{amssymb}
\usepackage{multirow}
\usepackage{tikz}
\usepackage{hyperref}

\usepackage[accepted]{icml2020}

\newcommand{\argmin}{\mathrm{argmin}} 

\def\R{\mathbb{R}}
\def\P{\mathrm{P}}

\def\our{HyperPocket}
\def\ours{HyperPocket(Rec)}

\icmltitlerunning{\our{}: Generative Point Cloud Completion}

\begin{document}

\twocolumn[
\icmltitle{\our{}: Generative Point Cloud Completion}



\icmlsetsymbol{equal}{*}

\begin{icmlauthorlist}
\icmlauthor{
Przemysław Spurek }{to}
\icmlauthor{Artur Kasymov}{to}
\icmlauthor{Marcin Mazur}{to}
\icmlauthor{Diana Janik}{to}
\icmlauthor{Sławomir Tadeja}{to}
\icmlauthor{Łukasz Struski}{to}
\icmlauthor{Jacek Tabor}{to}
\icmlauthor{Tomasz Trzciński}{goo}
\end{icmlauthorlist}

\icmlaffiliation{to}{Faculty of Mathematics and Computer Science, Jagiellonian University, Kraków, Poland}
\icmlaffiliation{goo}{Warsaw University of Technology, Warsaw, Poland}

\icmlcorrespondingauthor{Przemysław Spurek}{przemyslaw.spurek@uj.edu.pl}

\icmlkeywords{Machine Learning, ICML}

\vskip 0.3in
]



\printAffiliationsAndNotice{}  

\begin{abstract}

Scanning real-life scenes with modern registration devices typically give incomplete point cloud representations, mostly due to the limitations of the scanning process and 3D occlusions. Therefore, completing such partial representations remains a~fundamental challenge of many computer vision applications. 
Most of the existing approaches aim to solve this problem by learning to reconstruct individual 3D objects in a synthetic setup of an uncluttered environment, which is far from a real-life scenario. In this work, we reformulate the problem of point cloud completion into an {\it object hallucination task}. 
Thus, we introduce a~novel autoencoder-based architecture called \our{} that disentangles latent representations and, as a result, enables the generation of multiple variants of the completed 3D point clouds. We split point cloud processing into two disjoint data streams and leverage a hypernetwork paradigm to fill the spaces, dubbed {\it pockets}, that are left by the missing object parts. As a result, the generated point clouds are not only smooth but also plausible and geometrically consistent with the scene. Our method offers competitive performances to the other state-of-the-art models, and it enables a~plethora of novel applications.
\end{abstract}

\section{Introduction}
\label{submission}

The development of 3D registration devices, such as LIDARs or depth cameras, offers an unprecedented ability to capture complex 3D scenes and represent them with the so-called point clouds~\cite{yang2018pixor,kehoe2015survey}. One of the desired properties of this digitization process is the possibility to decompose real-life scenes into a set of distinct 3D objects so that they can be manipulated or modified, e.g., in the context of indoor redecoration or industrial facility reorganization. The main obstacles that prohibit currently used methods from solving this problem are limited sensor resolution, difficulty in covering all scanning positions,  
and 3D occlusions, resulting in incomplete point clouds~\cite{dai2017shape}.

Current approaches attempt to solve these problems by learning to complete point clouds of synthetic 3D objects~\cite{xie2020grnet,liu2020morphing,bello2020review}. Although the so-called point cloud completion algorithms allow for an effective increase of registration resolution, they struggle to reconstruct partially occluded objects, especially if they are part of complex indoor 3D scenes. This is mainly because they are trained to reconstruct a synthetically generated object, like those from the ShapeNet dataset~\cite{shapenet}, rather than completing or {\it hallucinating} 
the unseen part of it, as it is intuitively done by humans when
decomposing a complex scene containing occluded objects~\cite{nanay2018importance,briscoe2011mental}.
Furthermore, the reconstructions obtained by the classical point cloud completion methods can be inconsistent with the surrounding environment, as they do not take into account during the training the physical constraints of the scene or the scaling effect. Even when we leverage a multimodal shape completion framework~\cite{wu2020multimodal} with multiple versions of the objects generated, direct decomposition of the real-life scene remains infeasible, mainly due to the differences between the synthetic samples used to train the model and the outputs of the registered devices. 

\begin{figure*}[!h] 
\begin{center} 
 \includegraphics[width=\textwidth]{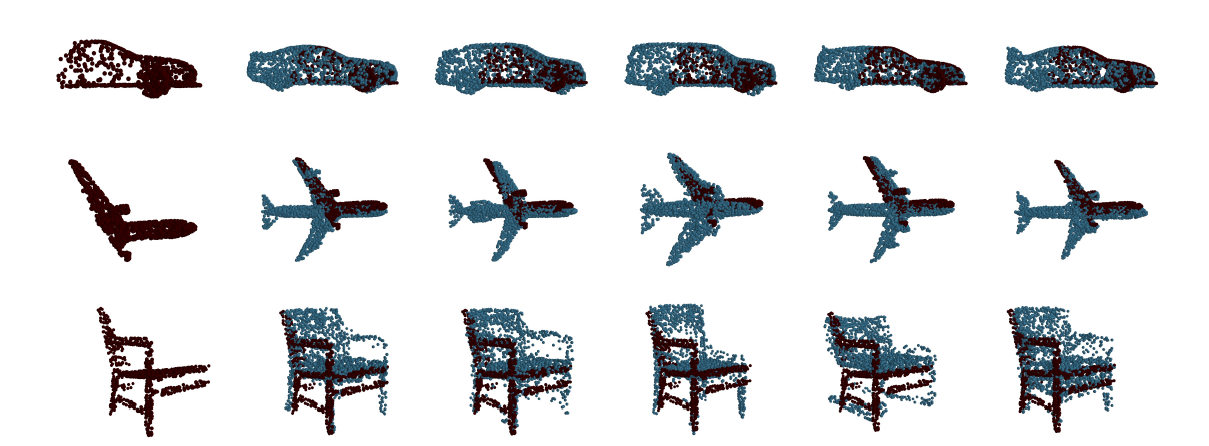}\\
\end{center} 
  \vspace{-0.5cm}
  \caption{Various 3D point clouds produced by our \our{} model. The first column shows an existing part of objects.
  In the next columns, we can see different versions of completions produced by our model.} 
\label{fig:generated_rec_point} 
\end{figure*}

In this work, we postulate to look at the problem of a 3D scene decomposition from a different perspective and 
reformulate the point cloud completion problem into an {\it object hallucination} task. This approach allows us to consider multiple ways to fill in a space occupied by a partially occluded 3D object and, hence, find a more plausible and geometrically coherent set of objects composing a given scene. 
For instance, if we only observe the frontal section of a given car, we can hallucinate a sedan, a hatchback, or a touring (wagon) tail of that car, as we show in Fig.~\ref{fig:generated_rec_point}. Following this intuition, we propose a novel completion point cloud method dubbed \our{} that aims at producing multiple variants of the complete point clouds. This naming emphasizes that the goal of our method is to {\it fill the pockets} left by the missing parts of an object and generate a realistic self-contained version of the object.

Our generative \our{} model leverages a hypernetwork approach to train an autoencoder that takes two complementary yet incomplete 3D point clouds and outputs weights of the so-called {\it target network}. 
This target network learns to map a probability distribution into a complete smooth point cloud of the 3D object, following the approach of~\cite{spurek2020hypernetwork,spurek2020hyperflow}. 
Hypernetworks can be understood as a meta-learning algorithm that produces weights to potentially smaller architectures. The main advantage of hypernetwork architecture in 3D point processing over the competing methods ({\it e.g.},~\cite{zamorski2018adversarial}) is that we can interpret target network weights as a representation of a smooth 3D distribution. Therefore we can produce any number of elements directly from the surface of the object.

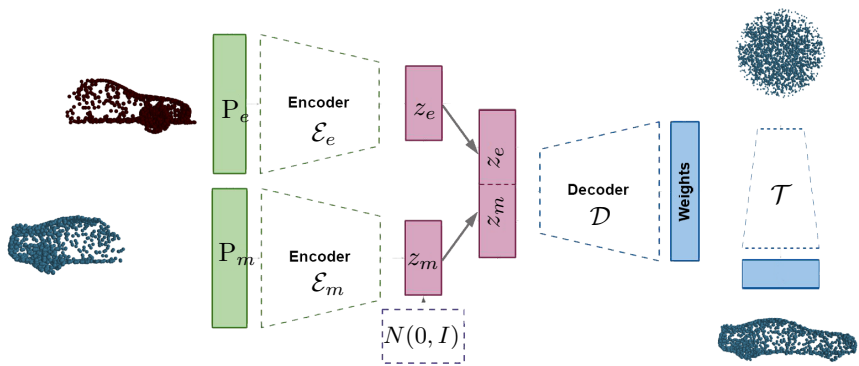
\begin{figure*}[t!] 
\begin{center} 
 \centering
    \begin{tikzpicture}[scale=1.]
    \node[inner sep=0pt] (russell) at (0,0)
    {\includegraphics[width=0.68\textwidth]{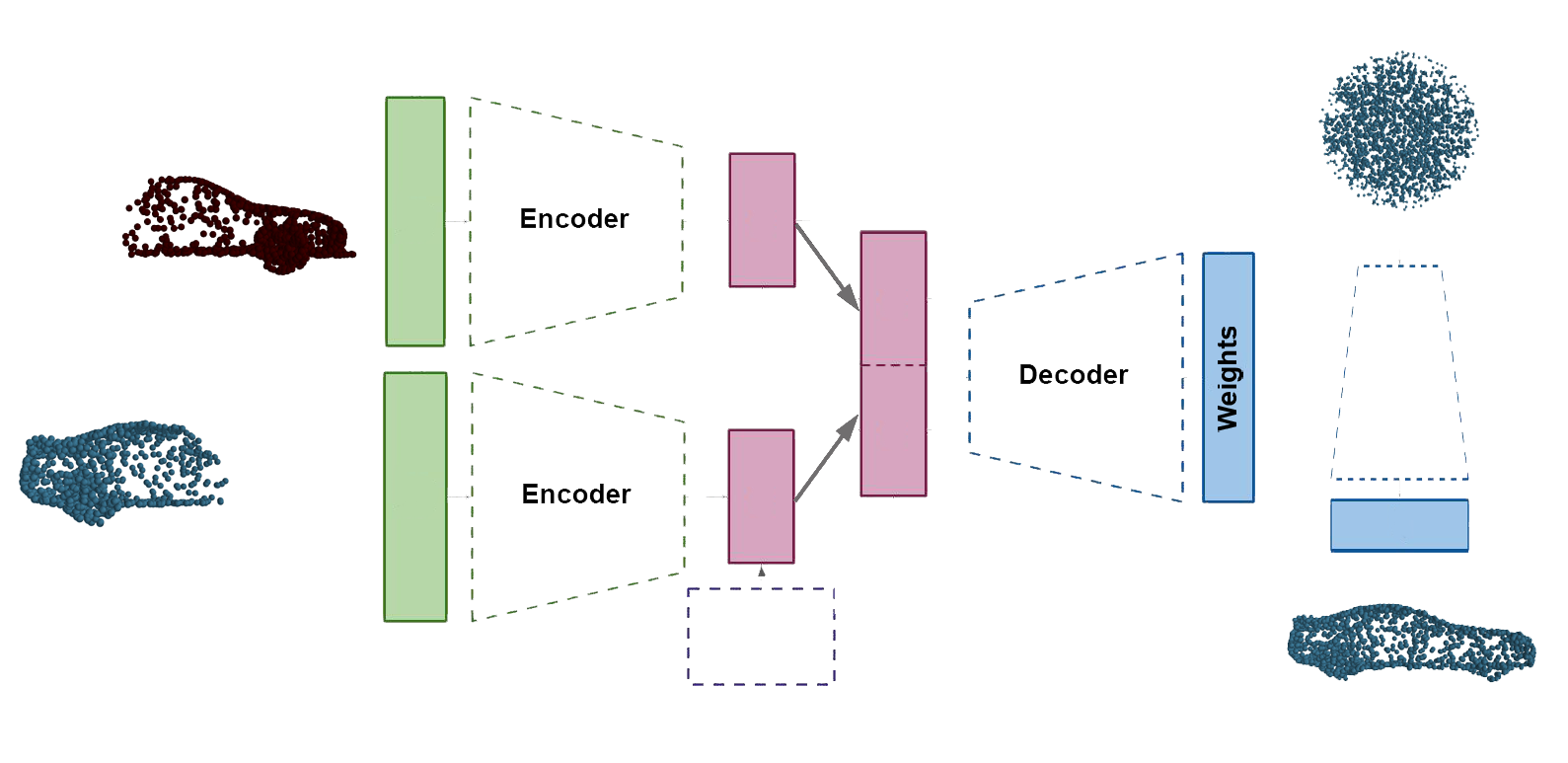} };
    \node[text width=3.5cm] at (0.1,0.8) {$ \mathcal{E}_e $};
    \node[text width=3.5cm] at (1.45,1.1) {$ z_e $};
    \node[text width=3.5cm] at (0.1,-1.2) {$ \mathcal{E}_m $};
    \node[text width=3.5cm] at (1.37,-0.9) {$ z_m $};
    
    \node[text width=3.5cm] at (3.8,-0.3) {$ \mathcal{D} $};    
    
    \node[text width=3.5cm] at (-1.1,1.1) {$ \P_e $};
    \node[text width=3.5cm] at (-1.14,-0.8) {$ \P_m $};
    \node[text width=3.5cm,rotate=90] at (0.8,1.3) {$ z_m \quad z_e $};
    
    \node[text width=2.0cm] at (0.3,-1.9) {\small $ N(0,I) $};
    \node[text width=3.5cm] at (6.2,0.) {$ \mathcal{T} $};
    
    \end{tikzpicture}
  \end{center}
  \vspace{-0.8cm}
  \caption{
Our \our{} model uses two encoders for producing separate latent representations for an existing part of an object and its missing part. By enforcing a Gaussian distribution on the latent space representing the missing part (the bottom part of the architecture), we can produce different point cloud completions. Then, both representations are concatenated and passed through a decoder, which outputs the weights of a target network used to produce a full 3D object.} 
 
\label{fig:teaser} 
\end{figure*}

The main intuition behind the proposed architecture is that by modeling latent representations of two separate encoder blocks, we train \our{} to produce two disentangled yet complementary object representations that can be modified separately, see Fig.~\ref{fig:teaser}. By enforcing the representation to follow a probability distribution, we can leverage the generative characteristics of the autoencoder. More specifically, when we decompose a 3D scene and hallucinate the missing parts of the occluded objects, we are able to sample distribution in the representation space and generate multiple variants of the completed point cloud. This formulation allows us to constrain our generations to follow the scene geometry, as shown in Fig.~\ref{fig:generated_ecene_point}, and adapt them accordingly, as displayed in Fig.~\ref{fig:adaptation_scene_1}. Our approach opens a plethora of new research paths and potential applications, {\it e.g.}, in virtual reality, gaming, and computer-aided generative design. 

Our contributions can be summarized as follows:
\vspace{-0.3cm}
\begin{itemize}
    \item we reformulate a point cloud completion problem into an object hallucination task inspired by how humans decompose complex 3D scenes with occluded objects into separate items, 
    \vspace{-0.3cm}
    \item we introduce a novel generative \our{} architecture based on an autoencoder designed with the hypernetwork paradigm, 
    \vspace{-0.3cm}
    \item our approach completes a real scene with objects that follow the scene geometry, while simultaneously adapting the generations to the existing environment.
    \vspace{-0.3cm}
\end{itemize}









\section{Related Work}

\paragraph{Shape Completion}
3D object completion is an important problem in computer vision with a multitude of deep learning methods proposed in the literature~\cite{xie2020grnet}. 
These methods typically use multilayer perceptrons (MLP), graphs, or convolutions as building blocks and various representations of point clouds to solve the completion problem.

Following the PointNet architecture  \cite{qi2017pointnet,qi2017pointnet++}, several works use a rotation and permutation invariant MLP for point cloud reconstruction \cite{yuan2018pcn,mandikal2019dense}. These methods model each point independently using several MLPs and then aggregate a global
feature using a symmetric function, such as max pooling. The state-of-the-art approaches AtlasNet \cite{groueix2018papier} and MSN \cite{liu2020morphing} recover the complete point cloud of an object by estimating a collection of parametric surface elements. 

By considering each point in a point cloud as a vertex of a graph, one can generate directed edges for a graph based on the neighbors of each point. Such an approach is used to create graph-based networks. Although the proposed solutions \cite{zhang2019linked,hassani2019unsupervised,wang2019deep} provide satisfactory reconstruction capabilities, they typically use additional information from a training dataset and cannot be directly applied in the real scene completion problem.

Another class of models uses a volumetric representation.
Early works \cite{dai2017shape,han2017high} apply 3D convolutional neural networks (CNNs) on voxels. Such solutions give satisfying results \cite{dai2017shape,mao2019interpolated,lei2019octree,li2018pointcnn} since such representation uses additional information from volumetric representation. In \cite{xie2020grnet}, authors propose a model that directly works on 3D points but uses voxelization as an element of architecture.

We compare our results with these and similar solutions in Section \ref{sec:rec}. We directly compare our results to AtlasNet \cite{groueix2018papier} since it is considered a current state-of-the-art model that uses only 3D point cloud representation and can be applied in 3D scene hallucination. 

\paragraph{Multimodal Shape Completion}

The most relevant to our approach is a Multimodal Shape Completion~\cite{wu2020multimodal} (MSC) method since it allows the generation of several multimodal outputs as a result of completing a single, partially visible 3D point cloud with a one-to-many mapping. 
This method uses three-stage training: The classical autoencoder and VAE are trained on a full 3D object in the first stage. Then it fixes the weights of models and produces latent representations of incomplete data as proposed in \cite{chen2019unpaired}. Lastly, a generative network completes the partial shape in a conditional generative modeling setting. The generation of the completions is conditioned on a learned mode distribution explicitly distilled from complete shapes. There are no paired completion instances in the training dataset. During the first stage, the model is trained only on the entire object, and during the last two steps, it uses only the partial ones. To construct a correct mapping between representations of full objects and partial ones the authors use GAN~\cite{goodfellow2014generative}.  

MSC approach yields a complex architecture that requires iterative training split into stages. 
In consequence, it is challenging to train the resulting model on the real-world data. Furthermore, as concluded by~\citet{wu2020multimodal}, this method shares the very same limitations as other approaches, particularly the inability to produce shapes with fine-scale details and the necessity for the input to be canonically oriented. In our work, we propose an end-to-end trained model that can be directly used in the case of the real-world scene (object hallucination task), see Fig.~\ref{fig:generated_ecene_point}. Moreover, we can also include geometrical constraints from the environment in our training and adapt the objects to follow them, as we show in Fig.~\ref{fig:adaptation_scene_1}.


\begin{figure*}[!h] 
\begin{center} 
 \includegraphics[height=3.4cm]{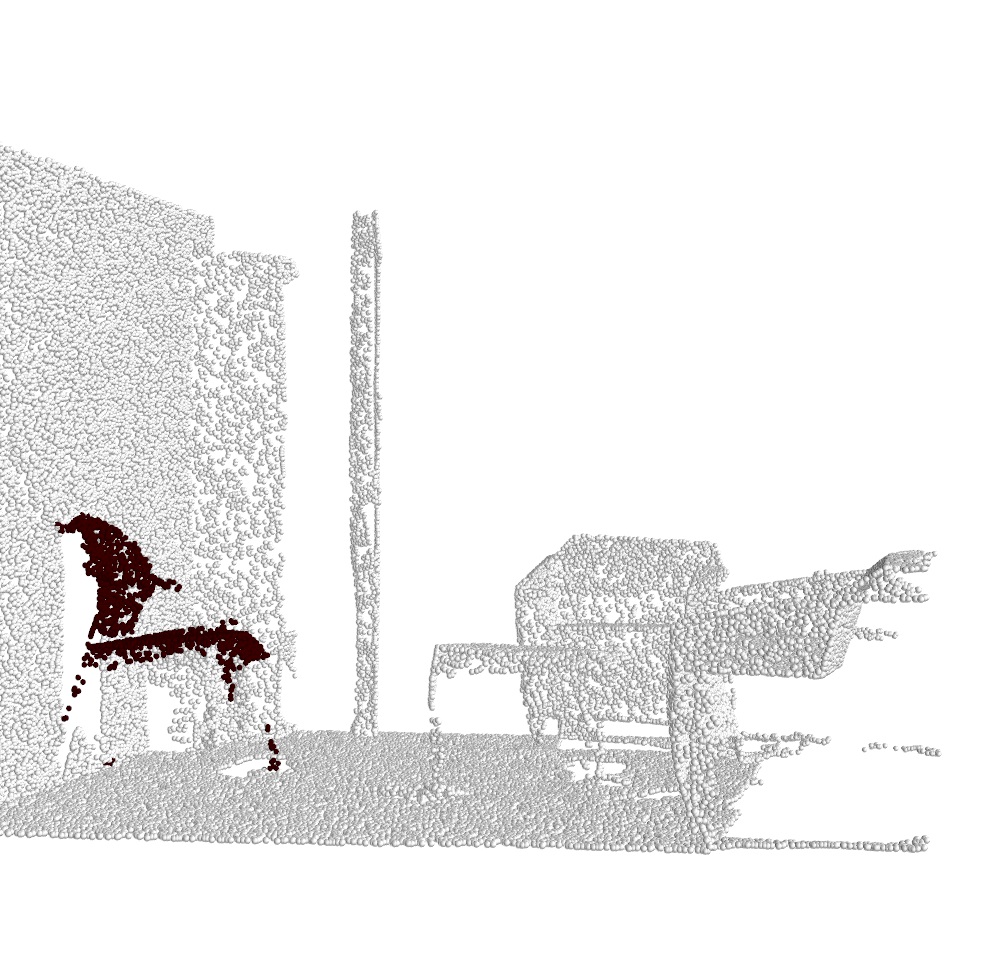} 
 \includegraphics[height=3.4cm]{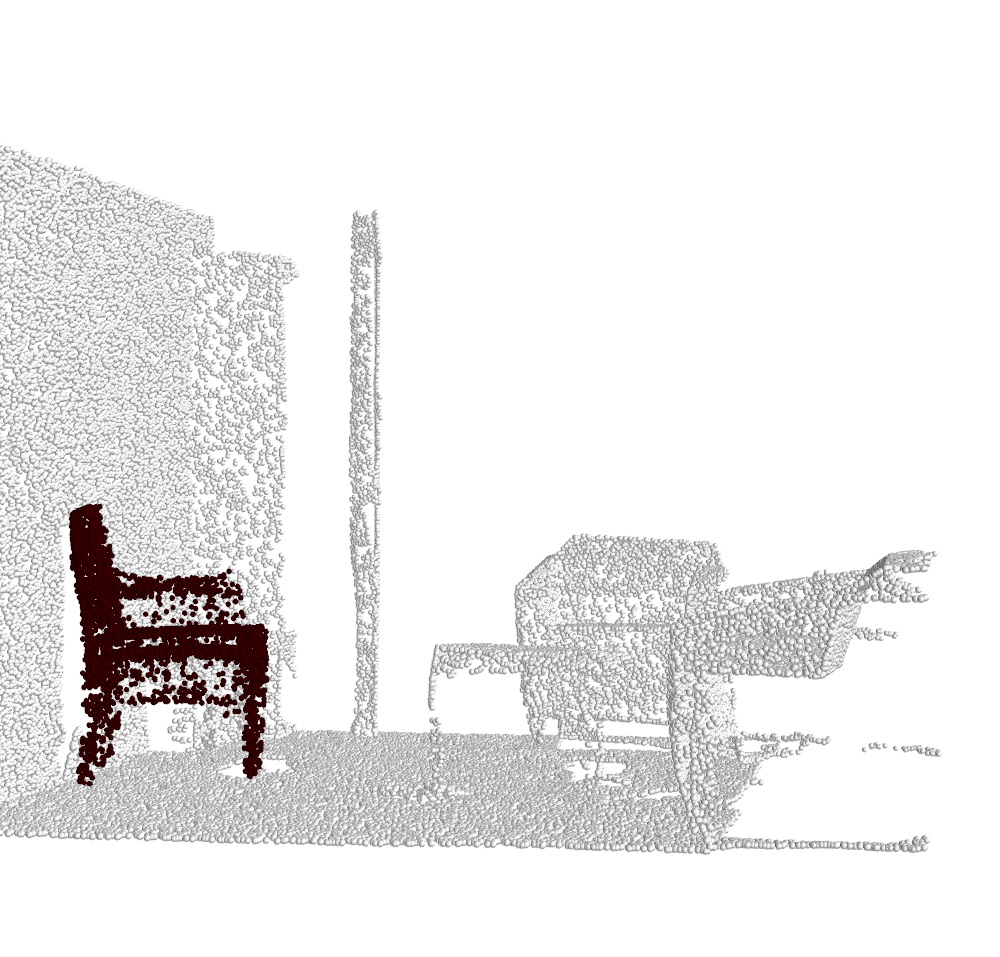} 
 \includegraphics[height=3.4cm]{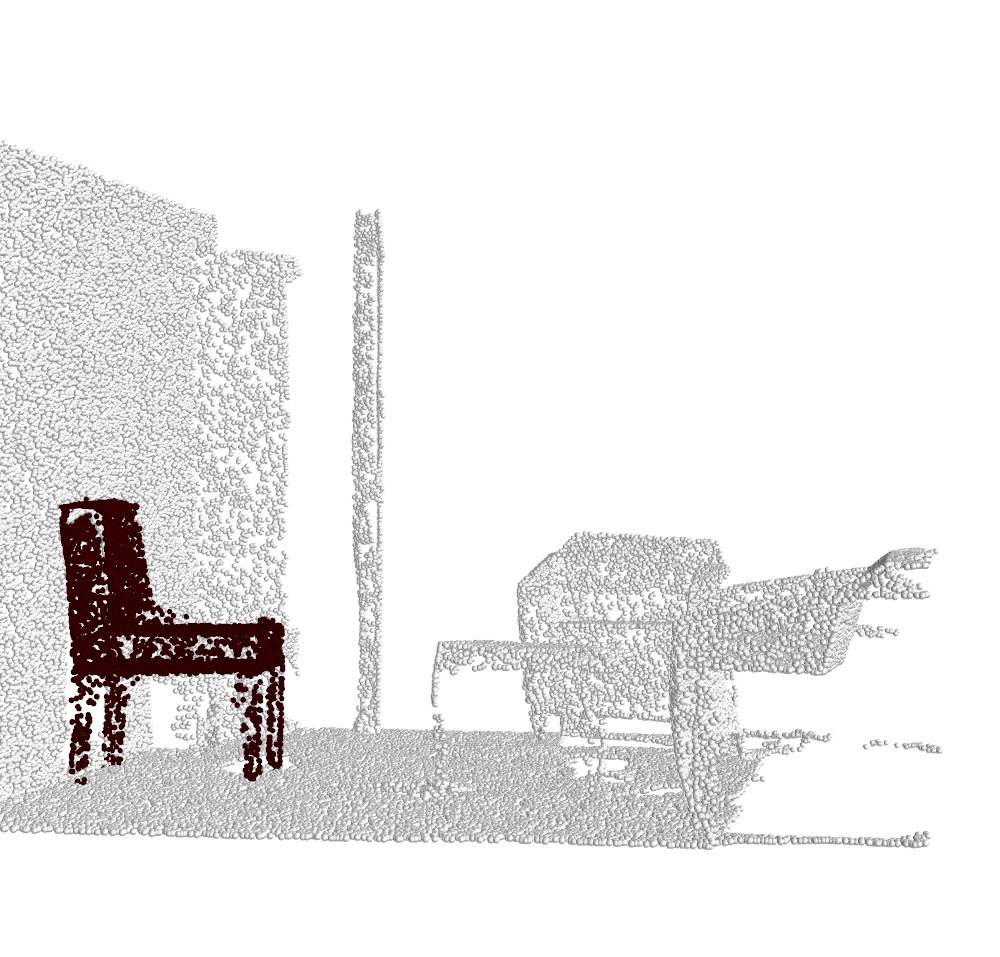}
 \includegraphics[height=3.4cm]{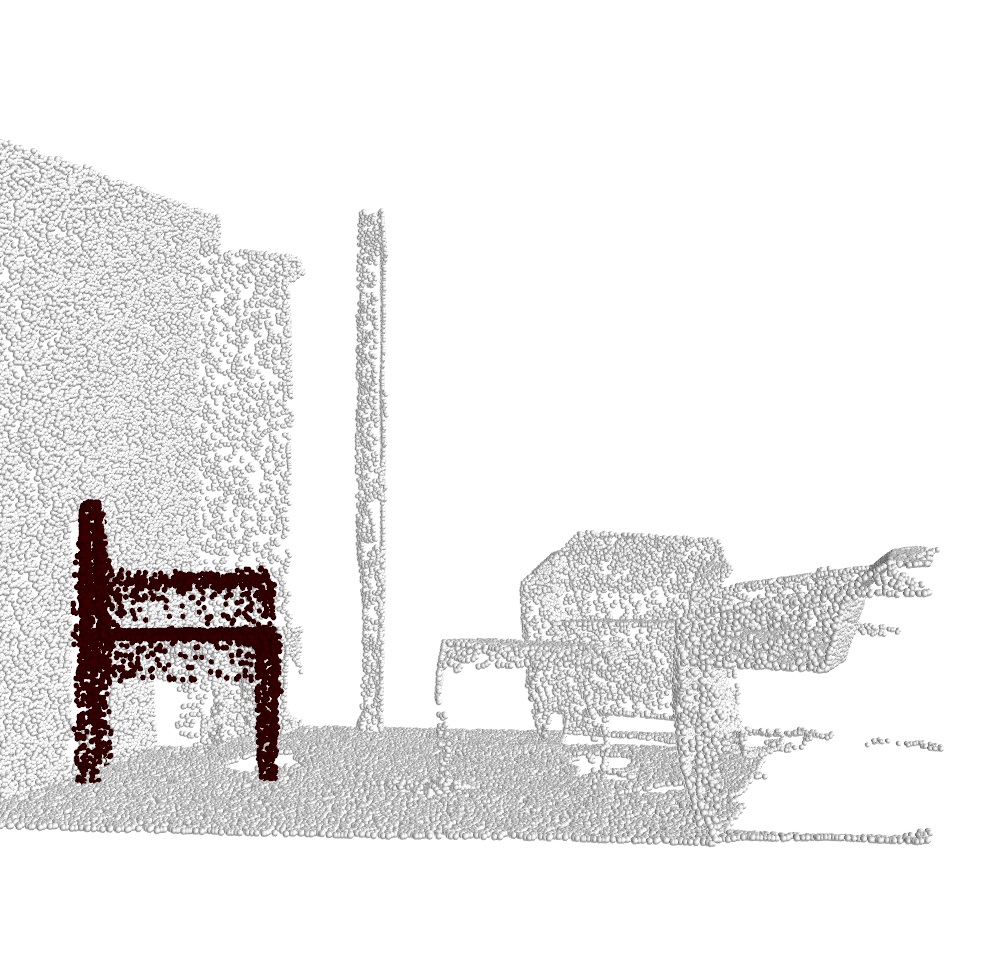} \\
 \includegraphics[height=3.4cm]{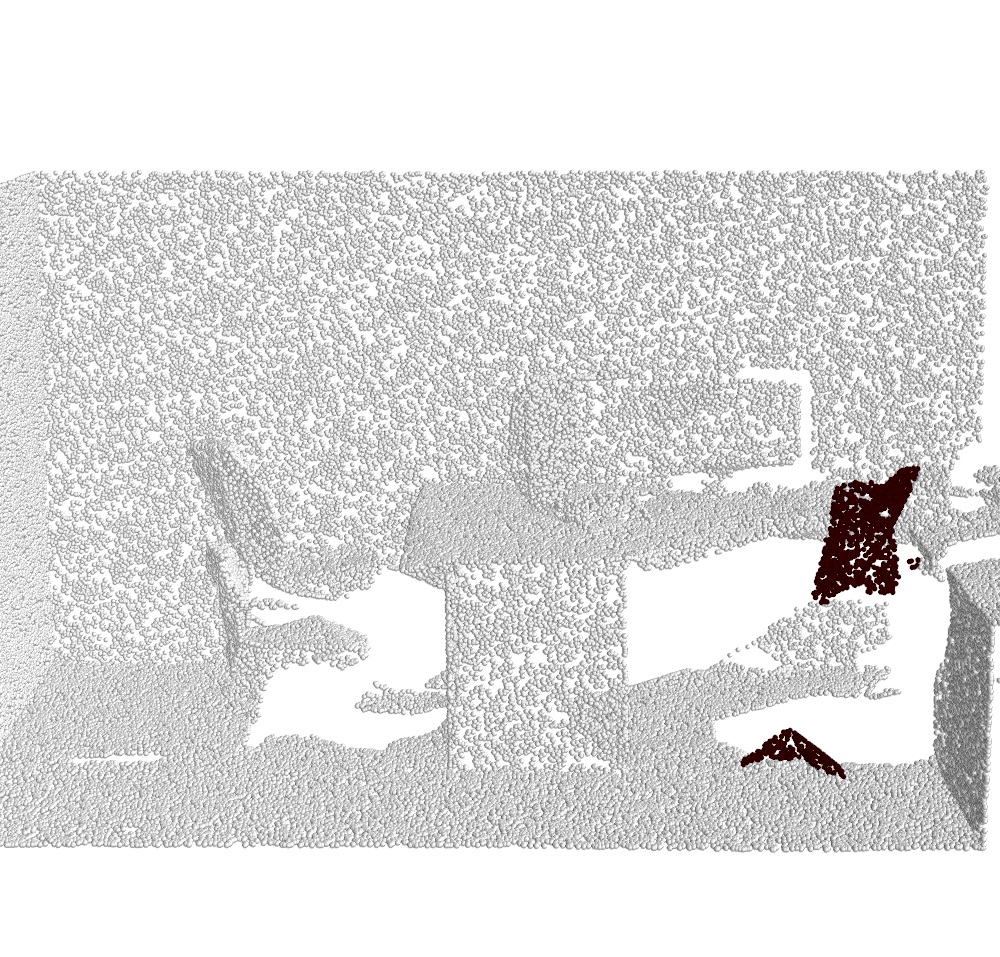} 
 \includegraphics[height=3.4cm]{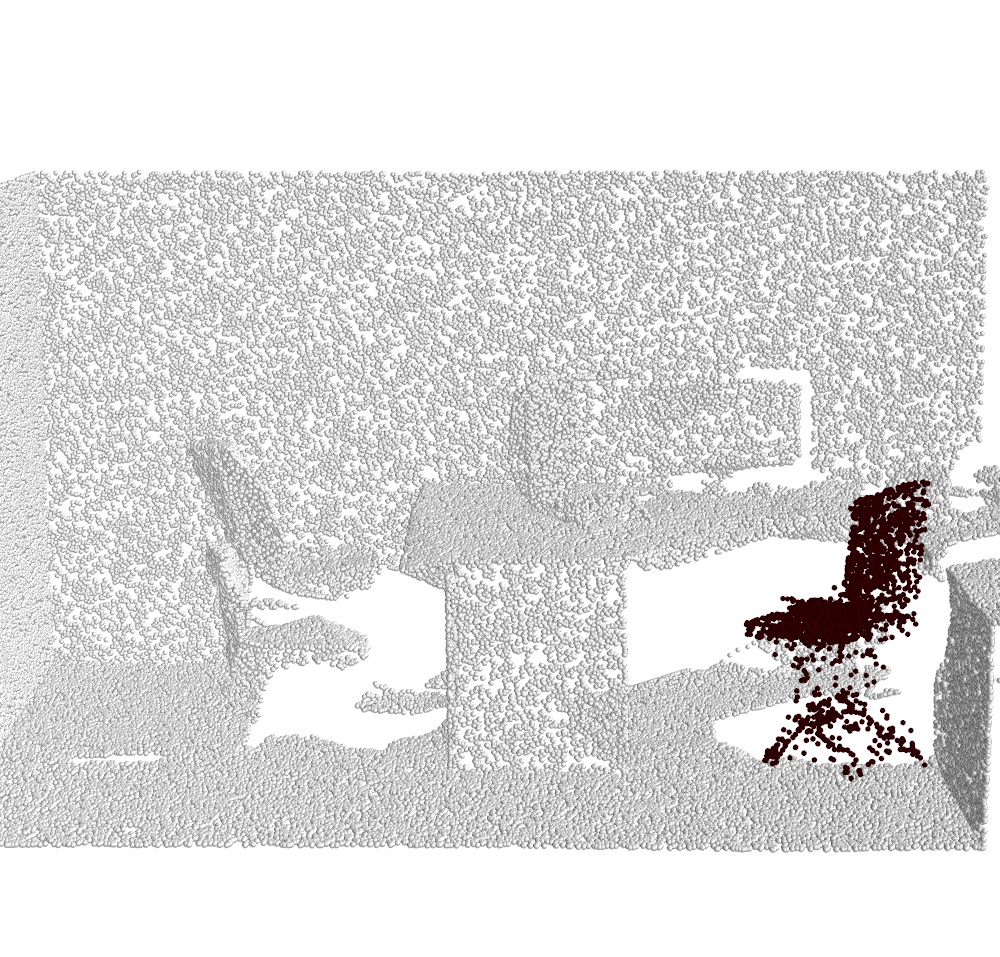} 
 \includegraphics[height=3.4cm]{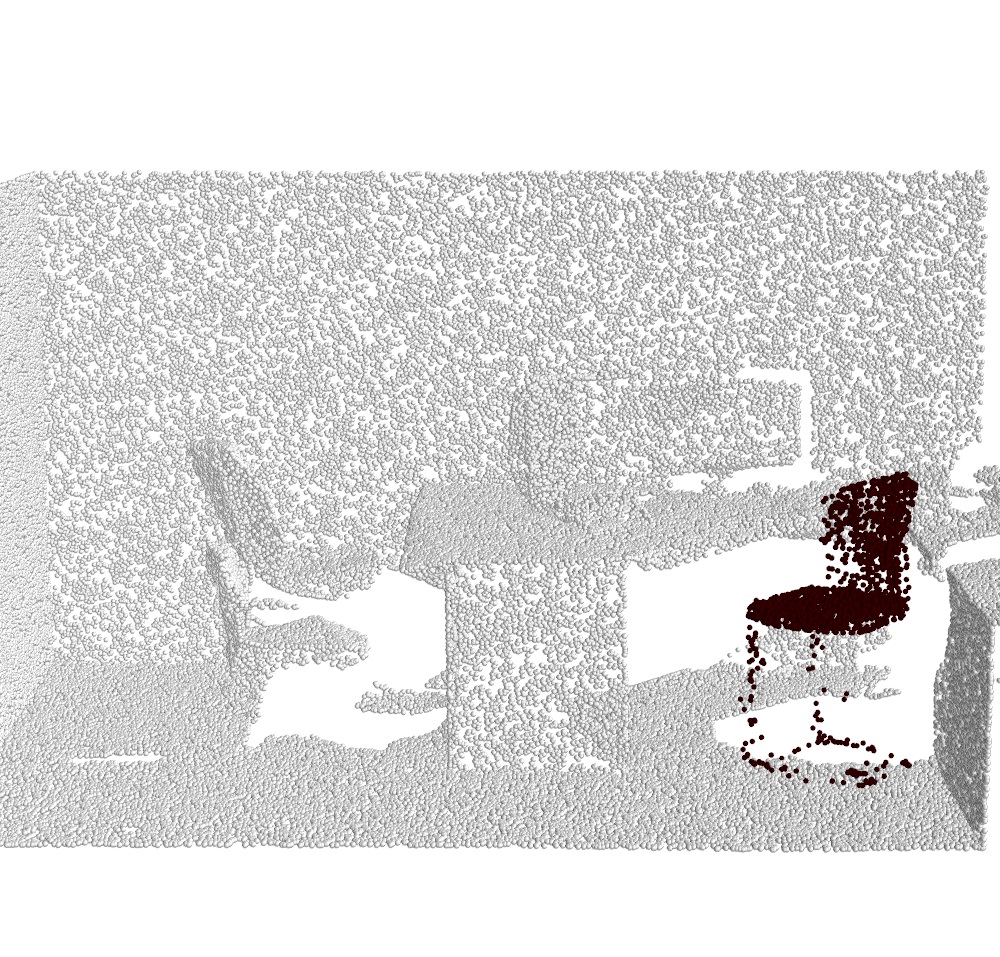}
 \includegraphics[height=3.4cm]{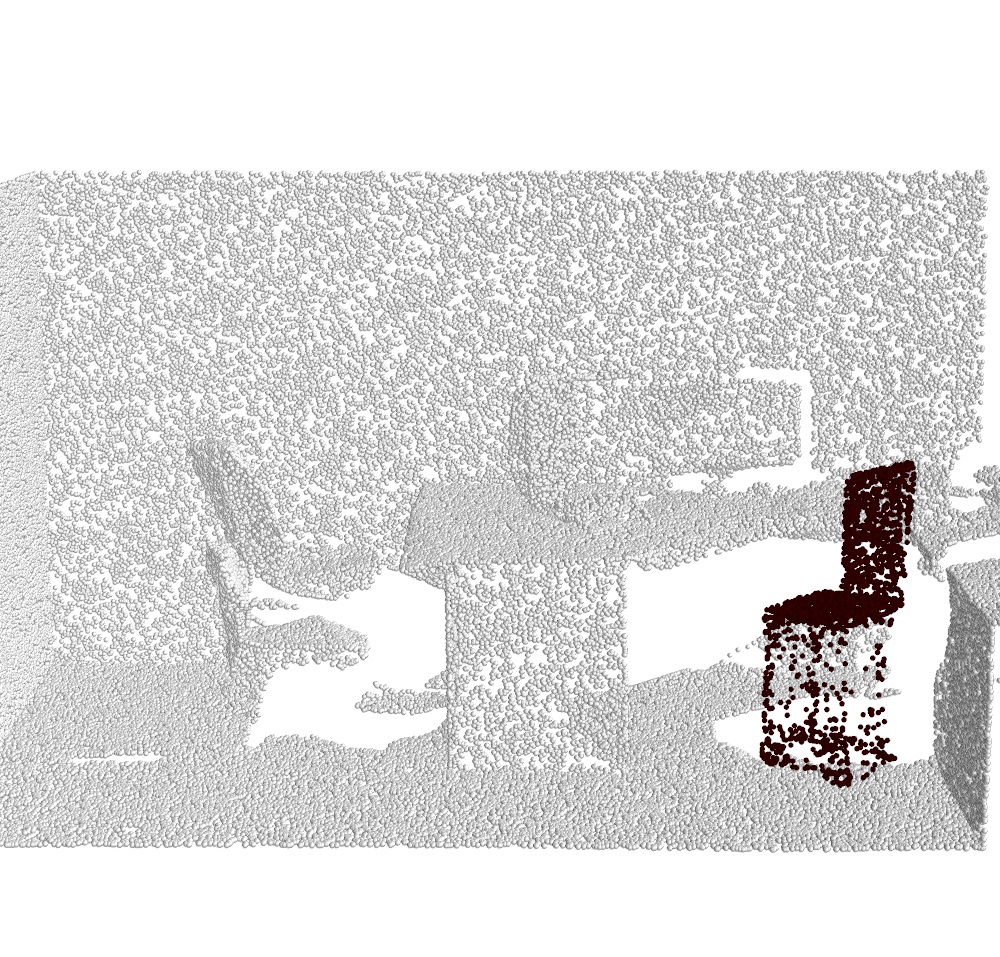} \\
\end{center} 
  \vspace{-0.1cm}
  \caption{Reconstructions of a chair in real-life scenes 
  produced by the
  \our{} model. In the first column, we present an original scene. In the next columns, we can see various possible hallucinations.} 
\label{fig:generated_ecene_point} 
\end{figure*}

\section{HyperPocket: a hypernetwork for 3D point cloud completion }
\label{sec:method}


In this section, we present the \our{} model designed to complete 3D point clouds. Before proceeding to the detailed description, we provide a brief idea of \our{}.

\vspace{-0.25cm}
\paragraph{\our{} main idea}
During training, we assume that we are given a splitting of the point cloud $\P$ into disjoint subsets $\P_e$ and $\P_m$, where $\P_e$
corresponds to the visible one, while $\P_m$ the missing one.
In the \our{} model we use two encoders for producing separate latent representations $z_e=\mathcal{E}_e(\P_e) \in \mathcal{Z}_e$ for an existing part of an object, and $z_m=\mathcal{E}_m (\P_m) \in \mathcal{Z}_m$ for a potentially missing one, see Fig.~\ref{fig:teaser}. We apply the decoder $\mathcal{D}$ on the concatenated representation $z=(z_e,z_m)$. By using the hypernetwork technique, the decoder is trained to build a distribution that is close to the point cloud $\P$.

However, the crucial novel aspect of \our{} is that we enforce the Gaussian distribution on the latent space $\mathcal{Z}_m$ corresponding to the missing part. Consequently, given a fixed visible part $\P_e$ of the point cloud $\P$,
the model can produce diversified completions of the incomplete point clouds by sampling $z \in \mathcal{Z}_m$ from the standard Gaussian distribution:
$
\mathcal{D}(\mathcal{E}_e\P_e,z) \mbox{ where }z \sim N_{Z_m}(0,I).
$
Therefore, using the generative aspects of \our{}, we can efficiently modify the produced reconstruction by changing $z \in Z_m$ to satisfy the desired constraints.

Before we present the further details, we briefly recall the main concepts lying behind our idea.

\vspace{0.1cm}
{ \bf Generative autoencoder model}
A classical autoencoder architecture is designed to represent high-dimensional data ({\it e.g.}, images) by their low-dimensional code vectors. Specifically, this architecture consists of an encoder
$\mathcal{E}\colon\mathcal{X}\longrightarrow{}\mathcal{Z}$
and a~decoder
$\mathcal{D}\colon\mathcal{Z}\longrightarrow{}\mathcal{X}$
networks, acting on a data space $\mathcal{X}$ and a latent (code) space $\mathcal{Z}=\mathbb{R}^d$, respectively, which are trained simultaneously to minimize distance between a given real data set $X=\{\P_1,\ldots, \P_n\}\subset \mathcal{X}$ and the set of reconstructions $\mathcal{D}(\mathcal{E}(X))=\{\mathcal{D}(\mathcal{E}(\P_1)),\ldots, \mathcal{D}(\mathcal{E}(\P_n))\}$. 
 We can also interpret this objective in the context of increasing the similarity between two probability distributions on $\mathcal{X}$ that represent the real and reconstructed data. Let us denote them as $P_X$ and $P_{\mathcal{D}(\mathcal{E}(X))}$, respectively. In order to obtain a generative framework, we should additionally ensure that the distribution of encoded data  $P_{\mathcal{E}(X)}$ is similar to a given prior (noise) distribution $P_Z$ (typically a Gaussian one) on the latent $\mathcal{Z}$.
 In the paper, we use Variational Autoencoders (VAE) \cite{kingma2013auto}. 

\vspace{0.1cm}
{ \bf Hypernetwork}
A hypernetwork architecture~\cite{ha2016hypernetworks} is a  neural network $\mathcal{H}$ that generates a vector of weights $\theta\in \Theta$ for a separate target network $\mathcal{T}_\theta$ designated to solve a specific task. 
The model is trained to learn $\mathcal{H}$ that returns an appropriate value of $\theta=\mathcal{H}(\P)$ for a given data point $\P\in \mathcal{X}$. Hypernetworks are widely used, {\it e.g.}, to generate a diverse set of target networks approximating the same function~\cite{sheikh2017stochastic}, as well as for the functional representations of images~\cite{klocek2019hypernetwork}.

\paragraph{HyperCloud}
HyperCloud~\cite{spurek2020hypernetwork} is a generative autoencoder-based model that uses a decoder to  
produce a vector of weights $\theta$ of a target network $\mathcal{T}_\theta\colon \mathbb{R}^3\to \mathbb{R}^3$. The target network is designed 
to transform a noise, {\it i.e.}, a sample from the uniform distribution on $S^2$ (the 2D unit sphere), into target objects (reconstructed data). 
In practice, we have one neural network architecture that uses different weights for each 3D object.
Note that here we describe the hypernetwork $\mathcal{H}={\mathcal{D}}\circ \mathcal{E}$.


\definecolor{Gray}{gray}{0.85}
\begin{table*}[t]
\caption{Point completion results on Completion3D compared using Chamfer Distance
($CD$). The value of $CD$ is computed on 2048 points and multiplied by $10^4$. Our \ours{} model gives results comparable to AtlasNet~\cite{groueix2018papier}.}
\label{rec_table}
\vskip 0.15in
\begin{center}
\begin{small}
\begin{sc}
\scalebox{0.90}{
\begin{tabular}{l|cccccccc|c}
\toprule
Method & Airplane & Cabinet & Car & Chair & Lamp & Sofa & Table & Watercraft & Overall\\
\midrule
GRNet          &   6.13 &  16.90 &   8.27 & \bf 12.23 & \bf 10.22 &  14.93 & \bf 10.08 & \bf  5.86 & \bf 10.64\\
SA-Net         &  \bf  5.27 & \bf  14.45 & \bf  7.78 & 13.67 & 13.53 & \bf 14.22 & 11.75 &  8.84 & 11.22\\
SoftPoolNet    &   \bf 5.27 & 18.66 & 10.32 & 15.56 & 12.80 & 14.92 & 11.99 &  6.74 & 12.11\\
TopNet         &  7.32 & 18.77 & 12.88 & 19.82 & 14.60 & 16.29 & 14.89 &  8.82 & 14.25\\
\cellcolor{Gray} AtlasNet       & \cellcolor{Gray} 10.36 & \cellcolor{Gray} 23.40 & \cellcolor{Gray} 13.40 & \cellcolor{Gray} 24.16 & \cellcolor{Gray} 20.24 & \cellcolor{Gray} 20.82 & \cellcolor{Gray} 17.52 & \cellcolor{Gray} 11.62 & \cellcolor{Gray} 17.77\\
\cellcolor{Gray} \ours{}  &  \cellcolor{Gray} 8.60 & \cellcolor{Gray} 25.34 & \cellcolor{Gray} 12.18 & \cellcolor{Gray} 25.08 & \cellcolor{Gray} 21.31 & \cellcolor{Gray} 19.81 & \cellcolor{Gray} 19.76 &  \cellcolor{Gray} 10.44 & \cellcolor{Gray} 17.91\\
PointSetVoting &  6.88 & 21.18 & 15.78 & 22.54 & 18.78 & 28.39 & 19.96 & 11.16 & 18.18\\
PCN            &  9.79 & 22.70 & 12.43 & 25.14 & 22.72 & 20.26 & 20.27 & 11.73 & 18.22\\
FoldingNet     & 12.83 & 23.01 & 14.88 & 25.69 & 21.79 & 21.31 & 20.71 & 11.51 & 19.07\\
\bottomrule
\end{tabular}
}
\end{sc}
\end{small}
\end{center}
\vskip -0.1in
\end{table*}


Training the HyperCloud model involves learning  $\mathcal{E}$ and  ${\mathcal{D}}$ to minimize the following objective function:\\
$$
\begin{array}{cc}
\!\! \mathcal{J}_{\mathrm{HC}} \! =\!\! \sum\limits_{i=1}^n \! DM(\P_i,\mathcal{T}_{\mathcal{H}(\P_i)}(u))\!+\! \lambda D_{KL}(P_{\mathcal{E}(X)},N(0,I)),
\end{array}
$$
where $X=\{\P_1,\ldots,\P_n\}$ is a given dataset, 
each $u_i$ is a set of points in $S^2$ sampled from the uniform distribution, and $\lambda$ is a hyperparameter that balances magnitude differences between the terms. Here $DM$ denotes 
a dissimilarity measure between 3D point clouds 
and $D_{KL}$ means the Kullback-Leibler divergence. 
Following \cite{spurek2020hypernetwork}, we can use as $DM$ Chamfer  Distance\footnote{One can also use Earth Mover’s (Wasserstein) distance ($EMD$).} ($CD$):
$$
\begin{array}{c}
CD(P,Q)=\sum\limits_{p \in P} \min\limits_{q\in Q}  \| p-q \|^2 + \sum\limits_{q \in Q} \min\limits_{p\in P}  \| p-q \|^2,
\end{array}
$$
where $\|\cdot \|$ denotes the Euclidean norm in $\mathbb{R}^3$.
\paragraph{\our{}}

Now we are able to recapitulate the construction of our \our{} model, which is designed to process partial data to produce multiple variants of a complete point cloud. The main modification we propose, for the vanilla autoencoder, is the introduction of a double encoder module, as displayed in Fig.~\ref{fig:teaser}. 
It involves dividing each input point cloud $\P\in \mathcal{X}$ into two disjoint subsets $\P_e$ and $\P_m$, representing existing and missing parts of an object, respectively. These subsets are then treated as inputs to the two independent encoders $\mathcal{E}_e$ and $\mathcal{E}_m$. 
The first of them ($\mathcal{E}_e$) processes $\P_e$ and is used only for reconstruction. The other ($\mathcal{E}_m$) works with $\P_m$ and renders a generative model since we force $z_m = \mathcal{E}_m(\P_m)$ to follow the Gaussian prior. At inference, we can sample data from the prior and obtain multiple variants of a completed object. 

To obtain a latent representation of a full object $z\in \mathcal{Z}$, we concatenate encoders' outputs $\mathcal{E}_e(\P_e)$ and $\mathcal{E}_m(\P_m)$. However, such a concatenation is not consistent since we model $z_e$ and $z_m$ independently. 
Therefore, we rely on a hypernetwork architecture to process the resulting representation and render a smooth and self-consistent object. More precisely, we train the decoder $\mathcal{D}$ to transfer such a representation into a vector of weights $\theta$ of a target network $\mathcal{T}_\theta\colon \mathbb{R}^3\to \mathbb{R}^3$. The target network 
transforms the noise on the unit sphere $S^2$ into target objects and outputs the final reconstruction. 

In practice, we optimize the following objective function:
$$
\mathcal{J} \!\! =\!\! \sum\limits_{i=1}^n DM(\P_i,\mathcal{T}_{\mathcal{H}(\P_i)}(u))\!\! + \!\! \lambda  D_{KL}(P_{\mathcal{E}_m(X_m)},N(0,I)),
$$
where  $\mathcal{H}(\P_i)=\mathcal{D}(\mathcal{E}_e(\P_{i,e}),
 \mathcal{E}_m(\P_{i,m}))$ for $i=1,\ldots, n$ and $X_m=\{\P_{1,m},\ldots,\P_{n,m}\}$.



We call the resulting model \our{}, which is inspired by the fact that the goal of our method is to fill in {\it the pocket}, left by the missing parts of an object, with multiple completion variants. 

\paragraph{\our{} adaptation to geometrical constraints}

Let us consider a more general situation, where we have
an already trained HyperPocket architecture, an existing
part of an object $\P_e$ and some external constraints given
by $C \geq  0$, e.g., following from a scene geometry. We assume that the minimal value of $C$ for a given point cloud means that it satisfies our constraints. So now we can restate the adaptation problem in the following manner: given $\P_e$ and $C$, find
a point $r$, potentially sampled from the Gaussian
prior, which minimizes the value of $C$. 
To be consistent with the existing part of the object, we also consider its distance from a target object, produced by $\mathcal{D}(\mathcal{E}_e(\P_e),r)$.
Thus, we arrive at the following
minimization problem:
$$
\argmin_{r} C\big( \mathcal{T}_{\mathcal{D}(\mathcal{E}_e(\P_e),r) }(u) \big) + CD\big( \mathcal{T}_{\mathcal{D}(\mathcal{E}_e(\P_e),r)}(u), \P_e\big),
$$
where $u$ is a set of points in $S^2$ sampled from the uniform distribution.
\begin{figure*}[!ht] 
\begin{center} 
 \includegraphics[width=0.9\textwidth]{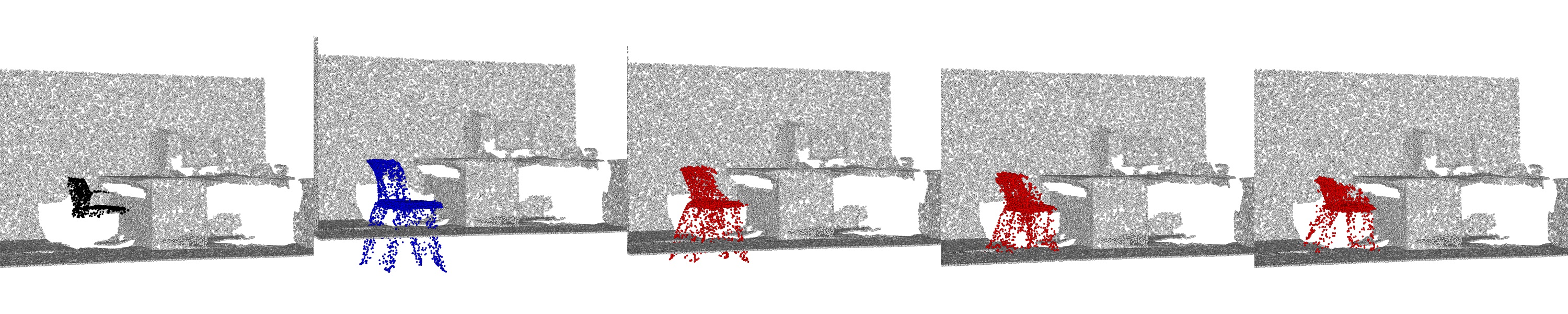}\\
   \vspace{-0.5cm}
 \includegraphics[width=0.9\textwidth]{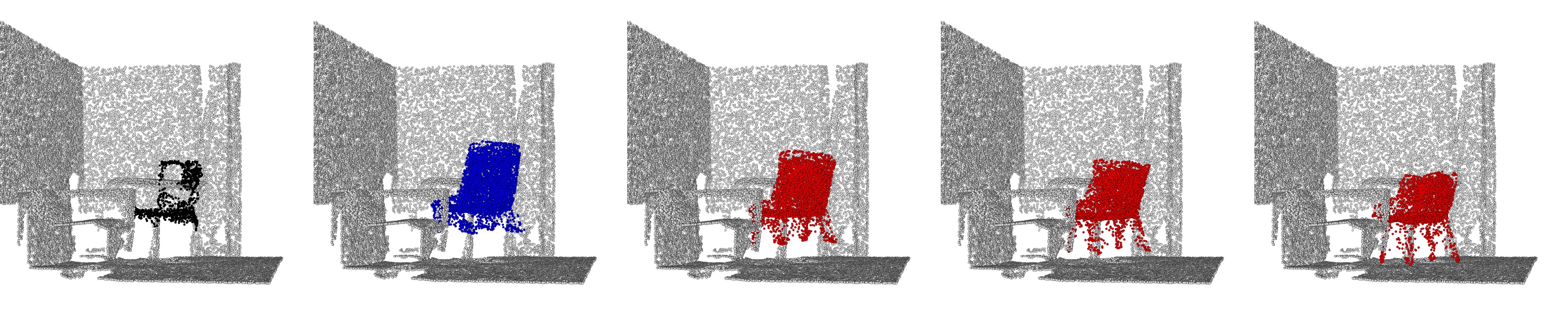}
\end{center} 
  \caption{Examples of chair reconstructions in scenes created by the \our{} model. The first column shows the scene with the incomplete chair (in black). The second column contains the initial reconstructions of the chair (in blue) created by the \our{} model. The next three columns present the process of adaptation.} 
\label{fig:adaptation_scene_1} 
\end{figure*}

\section{Experiments}
This section describes the experimental results of our model evaluated on the reconstruction and variant generation tasks. We follow the evaluation protocol of~\cite{yang2019pointflow} for the reconstruction tasks, as a typical setup for testing generative models does not include this scenario.
Therefore, we train our model using only the reconstruction function (we provide more details in the Appendix~A).
For the evaluation of generative capabilities, we follow two protocols to test the \our{} model. In the first one, we directly compare our model with \cite{wu2020multimodal}, as it is the most similar approach to ours. Although the results may not be directly comparable, given different data formats and processing setups, we adjust our training procedure to provide the most similar evaluation conditions. We also propose an additional testing scenario that evaluates the generative capabilities of models producing multiple completion variants. 
Finally, we present how the \our{} model works in the hallucination task for the real-world 3D scenes. We conclude this section with a discussion on the representation spaces learned by \our{}. 

\begin{figure}[!h] 
\begin{center} 
\includegraphics[height=3.5cm]{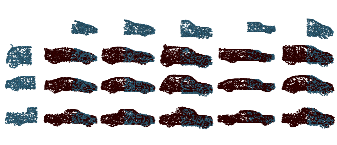}
\end{center} 
  \vspace{-0.5cm}
  \caption{Effect of reconstruction of only parts of objects. The first row presents the existing object part whereas the first column contains an input of a second encoder. The rest of the figure shows ``stitched'' reconstructions done by the \our{} model that inherit geometrical properties from inputs.} 
\label{fig:two_rec_point} 
\vspace{-0.4cm}
\end{figure}



\paragraph{Reconstruction capabilities}\label{sec:rec}
For the quantitative comparison of our method with the current state-of-the-art solutions in the reconstruction task, we follow the approach introduced in \cite{yang2019pointflow}. Here, the authors propose to compare the reconstruction capabilities of generative with non-generative models by training the model exclusively for the reconstruction task. Thus, without assumed prior on latent space. As such, we train the model only by using the reconstruction term of the objective function. We call this method \ours{}. The exact architecture of the \ours{} model is presented in Appendix~A. 
 
Reconstruction capabilities were verified against the Stanford Completion3D benchmark\footnote{\url{https://completion3d.stanford.edu}}. Using the model with the lowest Chamfer Distance ($CD$) on the validation set, we recover the complete point clouds for 1184 objects in the Completion3D testing set. According to the online leaderboard (see Tab.~\ref{rec_table}), the overall $CD$ for the \ours{} model is similar to the state-of-the-art AtlasNet model \cite{groueix2018papier} result and shows that our model produces compelling and useful reconstructions, even though it has not been designed to produce the exact object's reconstruction.
\paragraph{Generative capabilities}


To assess the generative capabilities of our model, we compare our results with the Multimodal Shape Completion (MSC) method introduced in \cite{wu2020multimodal}. However, the direct comparison of the HyperPocket to the MSC is rather unfair since the authors train the MSC model on data in different formats. Therefore, we also propose a new approach for evaluating such models.

\begin{table*}[t]
\caption{Quantitative comparison results on the 3D-EPN and PartNet \cite{wu2020multimodal} data sets. Note that
values of MMD (quality), TMD (diversity) and UHD (fidelity) 
are multiplied by $10^3$, $10^2$ and $10^2$, respectively.}
\label{gg_table}
\begin{center}
\begin{small}
\begin{sc}
\begin{tabular}{l|c@{$\:$}c@{$\:$}c@{$\:$}c|c@{$\:$}c@{$\:$}c@{$\:$}c|c@{$\:$}c@{$\ $}c@{$\ $}c}
\toprule
3D-EPN & \multicolumn{4}{c|}{MMD (lower is better) } &  \multicolumn{4}{c|}{TMD (higher is better) } & \multicolumn{4}{c}{UHD (lower is better)}\\
Method&Chair&Plane&Table&Avg.&Chair&Plane&Table& Avg.& Chair&Plane&Table&Avg.\\
\hline
pcl2pcl & 1.81 & 1.01 & 3.12 & 1.98 & 0.00 & 0.00 & 0.00 & 0.00 & \bf 5.31 & 9.71 & 9.03 & \bf 8.02 \\
KNN-latent & \bf 1.45 & 0.93 & \bf 2.25 & \bf 1.54 & 2.24 & 1.13 & 3.25 & 2.21 & 8.94 & \bf 9.54 & 12.70 & 10.40\\
MSC-im-l2z & 1.91 & 0.86 & 2.78 & 1.80 & 3.84 & 2.17 &  4.27 & 3.43 & 9.53 & 10.60 & 9.36 & 9.83 \\
MSC-im-pc2z & 1.61 & 0.91 & 3.19 & 1.90 & 1.51 & 0.82 & 1.67 & 1.33 & 8.18 & 9.55 & \bf 8.50 &  8.74 \\
MSC & 1.61 & \bf 0.82 & 2.57 & 1.67 & 2.56 & 2.03 & 4.49 &  3.03 & 8.33 & 9.59 & 9.03 & 8.98\\
\our{} & 2.61  & 1.39 & 5.15 & 3.05 & \bf 4.15 & \bf3.29  & \bf 5.47 & \bf 4.30 & 11.85 & 13.59 & 14.28 & 13.24 \\
\end{tabular}

\begin{tabular}{l|c@{$\:$}c@{$\:$}c@{$\:$}c|c@{$\:$}c@{$\:$}c@{$\:$}c|c@{$\:$}c@{$\ $}c@{$\ $}c}
\toprule
PartNet & \multicolumn{4}{c|}{ MMD (lower is better) } &  \multicolumn{4}{c|}{ TMD (higher is better) } & \multicolumn{4}{c}{UHD (lower is better)}\\
Method & Chair & Lamp & Table & Avg. & Chair & Lamp & Table & Avg. & Chair & Lamp & Table & Avg.\\
\hline
pcl2pcl & 1.90 &  2.50 & 1.90 &  2.10 & 0.00 & 0.00 & 0.00 & 0.00 & \bf 4.88 & \bf 4.64 & \bf  4.78  & \bf 4.77 \\
KNN-latent & \bf 1.39 & \bf 1.72 & \bf 1.30 & \bf 1.47 & 2.28 & 4.18 & 2.36 & 2.94 & 8.58 & 8.46 & 7.60 & 8.22\\
MSC-im-l2z & 1.74 & 2.36 & 1.68 & 1.93 &  3.74 &  2.68 &  \bf 3.59 & 3.34 & 8.41 & 6.37 & 7.21 & 7.33 \\
MSC-im-pc2z & 1.90 & 2.55 & 1.54 & 2.00 & 1.01 & 0.56 & 0.51 & 0.69 & 6.65 & 5.40 & 5.38 & 5.81 \\
MSC & 1.52 &  1.97 & 1.46 & 1.65 & 2.75 & 3.31 & 3.30 &  3.12 & 6.89 & 5.72 & 5.56 & 6.06 \\
\our{}  & 2.46 &  5.05  & 4.88 & 4.13 & \bf 14.75 & \bf 26.43 & 0.93 & \bf 14.04 & 14.76  & 17.28 & 12.25 & 14.76\\
\end{tabular}

\end{sc}
\end{small}
\end{center}
\end{table*}


Following the evaluation approach proposed in \cite{wu2020multimodal}, we use only three object categories from each of the 3D-EPN \cite{dai2017shape} and PartNet \cite{Mo_2019_CVPR} datasets.
For each partial shape in the test set, we generate $k = 10$ completion results and adopt these measures for quantitative evaluation: Minimal Matching Distance ($MMD$), Total Mutual Difference ($TMD$), and Unidirectional Hausdorff Distance ($UHD$). The results of this comparison are presented in Tab.~\ref{gg_table}. As we can see, in general, \our{} obtains comparable results to other approaches. However, one can observe that \our{} obtains worse values of $MMD$ and $UHD$ measures, which describe the quality and reconstruction level of the completed shape.
The reason behind that is that, contrary to other models, \our{} builds completely split representations 
for parts of the model (no part ``sees'' the whole input point cloud). The dual reasoning explains why \our{} obtains a much better diversity, which is measured by $TMD$ factor. These results emphasize the fundamental advantage of our method, which is the ability to effortlessly and efficiently generate any number of similar, plausible 3D objects that naturally complete the given scene.
\paragraph{New evaluation framework for generative models}

We propose a new method to evaluate the point cloud generation that couples the existing metrics using PointFlow~\cite{yang2019pointflow}. Since the datasets for the reconstruction task are provided differently, we have prepared a new ShapeNet version called MissingShapeNet\footnote{Data and the code for evaluation is available on \url{https://github.com/gmum/3d-point-clouds-autocomplete} 
}.

For a fair comparison, we take the same train/val/test split as in PCN \cite{yang2018foldingnet}.
For elements in the train and validation sets, we have made four different divisions into equal existing and missing parts using random 2D planes. Since we need to verify how the missing part is being generated, we need to have always the same partition of each object. Hence, the test set elements we divide into left and right parts. 


We calculate Jensen-Shannon Divergence ($JSD$), Coverage ($COV$), Minimum Matching Distance ($MMD$) measures, and we average the results across all elements in the test data sets (see~Table~\ref{tab:gen_results}). For a more detailed description of measures see Appendix~B. 
As we can see, our \our{} model gives much better results than the classic approach relying on generating only one version of the object. Such a result confirms that \our{} is a generative model.

\paragraph{Adaptation to a real scene}

In the context of a real scene completion task, we train our model with rotations around the vertical axis as part of an augmentation procedure. Thanks to such a solution, we can produce samples with different rotations (see Fig.~\ref{fig:generated_ecene_point}). 
Such procedure works very well in the case of a scene where the existing part of an object contains all boundary points from the ground truth object.
For instance, in the case of chairs (see Fig.~\ref{fig:adaptation_scene_1}), our model needs to be informed about the scene's orientation, e.g., it has to be aware of where the scene's floor is, to, for instance, reconstruct the correct leg length. When we have only used the chair \our{} produces elements with different leg lengths. Consequently, the object does not fit properly into the scene's environment.
\begin{table}[]
\begin{center}
\caption{Generation results for the \ours{} and \our{} models. MMD-CD scores are multiplied by
$10^3$, while MMD-EMD and JSDs by $10^2$. }
\label{tab:gen_results}
\scalebox{0.7}{
\begin{sc}
\begin{tabular}{l@{$\:$}l@{$\:$}|@{$\:$}c@{}cccc}
\hline
\multirow{2}{*}{Category} & \multirow{2}{*}{Methods} & \multirow{2}{*}{JSD ($\downarrow$)} & \multicolumn{2}{c}{MMD($\downarrow$)} & \multicolumn{2}{c}{COV($\%$, $\uparrow$)}  \\  
                           &                          &                      & CD          & EMD       & CD         & EMD                \\ \hline
\multirow{2}{*}{Airplane} & \ours{} &  3.64 &  4.33 &  8.87 &  1.18 &  3.61 \\
 & \our{}  & \bf 2.87 & \bf 2.53 & \bf 7.69 & \bf 10.33 & \bf 16.06 \\ \hline
\multirow{2}{*}{Cabinet} & \ours{} &  5.93 &  17.01 &  14.44 &  0.92 &  2.14 \\
 & \our{}  & \bf 4.33 & \bf 10.92 & \bf 13.03 & \bf 4.93 & \bf 6.95 \\ \hline
\multirow{2}{*}{Car} & \ours{} &  2.74 &  2.05 &  6.65 &  1.19 &  5.26 \\
 & \our{}  & \bf 1.65 & \bf 1.48 & \bf 6.42 & \bf 11.16 & \bf 10.95 \\ \hline
\multirow{2}{*}{Chair} & \ours{} &  5.55 &  12.01 &  13.94 &  0.73 &  1.69 \\
 & \our{}  & \bf 3.39 & \bf 6.74 & \bf 11.25 & \bf 5.90 & \bf 7.20 \\ \hline
\multirow{2}{*}{Lamp} & \ours{} &  6.68 &  30.96 &  25.31 &  0.76 &  1.35 \\
 & \our{}  & \bf 5.96 & \bf 19.49 & \bf 21.27 & \bf 4.88 & \bf 5.27 \\ \hline
\multirow{2}{*}{Sofa} & \ours{} &  5.02 &  8.66 &  11.54 &  0.78 &  1.78 \\
 & \our{}  & \bf 2.98 & \bf 5.58 & \bf 9.85 & \bf 7.56 & \bf 8.00 \\ \hline
\multirow{2}{*}{Table} & \ours{} &  6.21 &  19.30 &  16.02 &  0.79 &  1.41 \\
 & \our{}  & \bf 4.30 & \bf 11.33 & \bf 12.90 & \bf 4.54 & \bf 5.80 \\ \hline
\multirow{2}{*}{Watercraft} & \ours{} &  4.80 &  11.77 &  13.15 &  0.96 &  2.24 \\
 & \our{}  & \bf 3.52 & \bf 6.77 & \bf 11.52 & \bf 8.90 & \bf 8.38 \\ \hline
\multirow{2}{*}{Average} & \ours{} &  5.07 &  13.26 &  13.74 &  0.91 &  2.44 \\
 & \our{} & \bf 3.62 & \bf 8.11 & \bf 11.74 & \bf 7.28 & \bf 8.58 \\ \hline
 
\end{tabular}
\end{sc}
}
\end{center}
\end{table}

\begin{figure}[t!] 
\begin{center} 
 \includegraphics[height=2.9cm]{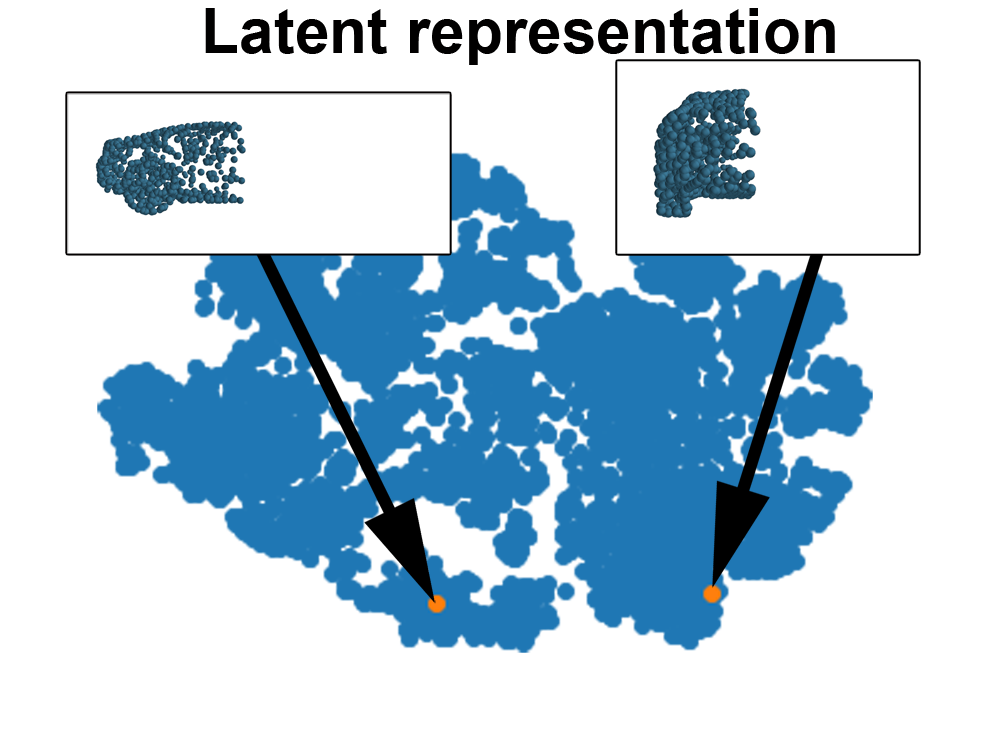} \
 \includegraphics[height=2.9cm]{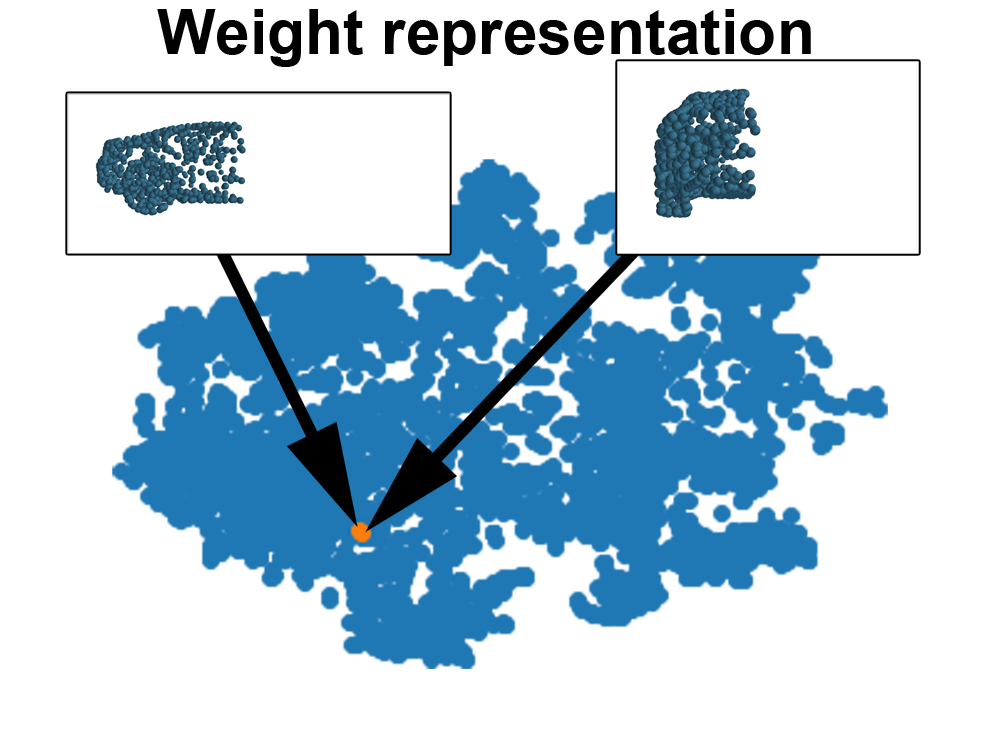}\\ 
 \includegraphics[height=2.9cm]{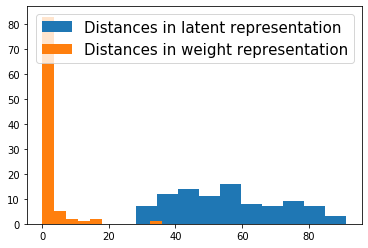} 
\end{center}  
\vspace{-0.3cm}
  \caption{A t-SNE plot for the latent and weight space obtained from the \our{} model (car category). The embeddings of a single car taken from a different point of view in the latent space are different, but we obtain similar representations in weight space. Latent space encodes each part of the object separately, and the target network merges it into a single representation. }
  \label{fig:latent}
\end{figure}

Contrary to MSC \cite{wu2020multimodal}, our model is trained in an end-to-end manner and can be adapt to the scene reconstruction task. For this purpose, we use a pre-trained \our{}  with frozen weights. 
In the first step, we take some partial data $\P_e$ (see the first column of Fig.~\ref{fig:adaptation_scene_1}). Next, we produce a latent representation of the element $\mathcal{E}_e(\P_e)$. Then, to produce some realization of the missing part, we sample $r$ from Gaussian distribution. Now our first version of a full object is given by the target network with weights ${\mathcal{D}}(\mathcal{E}_e(\P_e), r)$ (see the second column of Fig.~\ref{fig:adaptation_scene_1}).

To ensure correct scaling and fitting of the hallucinated objects into the scene, we impose certain constraints on the model. For example, in the case of chairs, first of all, we need them to be situated on the floor. The distance between object and floor should be zero. On the other hand, we add distance between a final object and the existing part of the object to be consistent with the reconstruction task. Now we make $z_m$ as a trainable parameter and minimize the new cost function:
$$
\min_{r} CD( \! \mathcal{T}_{{\mathcal{D}}(\mathcal{E}_e(\P_{\!e}), r)}(u) ,  \P_{\!e} )
\! + \! CD(\! \mathcal{T}_{{\mathcal{D}}(\mathcal{E}_e(P_e), r))}(u) , \! \P_{\!f} ),
$$
where $\P_f$ is a part of floor and $u$ is a set of points sampled from the uniform distribution on $S^2$.
In Fig.~\ref{fig:adaptation_scene_1}, we show few steps of the adaptation procedure.

\vspace{-0.3cm}
\paragraph{Representation learning}


The idea described in this paper is to model latent representation in two blocks dedicated independently to the existing and missing part of an object (see Fig.~\ref{fig:teaser}). Such two individual representations are merged into a single self-consistent and smooth 3D object. To achieve this, we use two encoders, one for each part of an object. Then, these representations are concatenated into a single one. Next, the decoder transfers such representation into the weight of the target network.


Our approach for each 3D point cloud produces two representations: classical latent code and weights of the target network. Such representations are in some sense similar since weights of the target network are a transformation of latent codes by a decoder. However,  they have different structures. The latent contains disentangled representation produced by autoencoder architecture. This representation consists of a separate block with which we can model independently two parts of a given object. On the other hand, weights of the target network represent operations that we must apply to transform the sample from prior distribution into a 3D object. Such representation is obtained from latent codes by the decoder and can be understood as a formula to produce an object which has properties encoded in latent representation.

Consequently, these two representations encode completely different information, despite the imaginary similarity. To verify it, we take already trained \our{} and visualize both representations, as shown in Fig.~\ref{fig:latent}. First, we take one class (e.g., cars), construct its latent and weight representations, and use t-SNE to visualize the results. In Fig.~\ref{fig:latent}, we present a representation of the car class. Then we take a single object and take partial data from different positions. We split the car in two ways, i.e., we slice along and across the model. 
In Fig.~\ref{fig:latent}, we present the embedding of such two positions in latent space and space of weights. As we can see, the same object with two different points of view is represented by completely different vectors in the latent space. Latent space encodes separately and independently each part of an object. In the case of weight representation, we obtain almost similar vectors. In Fig.~\ref{fig:latent}, we show a histogram of distances for one hundred randomly chosen objects from the test set. As we can see, such representation is invariant to the point of view.

Moreover, we show that \our{} can reconstruct an object from parts taken from two different objects within the same class. An example of this can be seen in Fig.~\ref{fig:two_rec_point} where we present combined reconstructions of various fronts (columns) and rears of cars (rows). As we can see, the \our{} model constructs plausible car shapes that inherit geometrical properties from the input data, and the target network correctly ``stitches'' such parts.

\section{Conclusion}

This paper presents a new look at a point cloud completion problem and considers its generative extension to the object hallucination task. Our goal is to produce multiple possibilities of completing a given point cloud instead of providing a single reconstruction. To solve such a problem, we proposed a novel generative \our{} architecture based on an autoencoder designed using the hypernetwork paradigm. Such a solution can produce many different completion outputs hallucinating unseen parts of objects in a scene, as intuitively done by humans when decomposing a complex scene containing occluded objects. We show that our method provides competitive performances to the state-of-the-art approaches and can be successfully applied to real-world scenes.

 \clearpage

\section{Appendix~A: Architecture of \ours{}}

For the quantitative comparison of our method with the current state-of-the-art solutions in the reconstruction task, we follow the approach introduced in \cite{yang2019pointflow}. The authors propose to compare the reconstruction capabilities of generative with non-generative models by training the model exclusively for the reconstruction task, without assumed prior on latent space. Thus, we train the model only by using the reconstruction term of the objective function. We call this method \ours{}.

In \ours{} we use only one latent part without the generative part, see Fig.~\ref{fig:generated_ecene_point}  

\begin{figure}[!h] 
\begin{center} 
 \includegraphics[height=3.5cm]{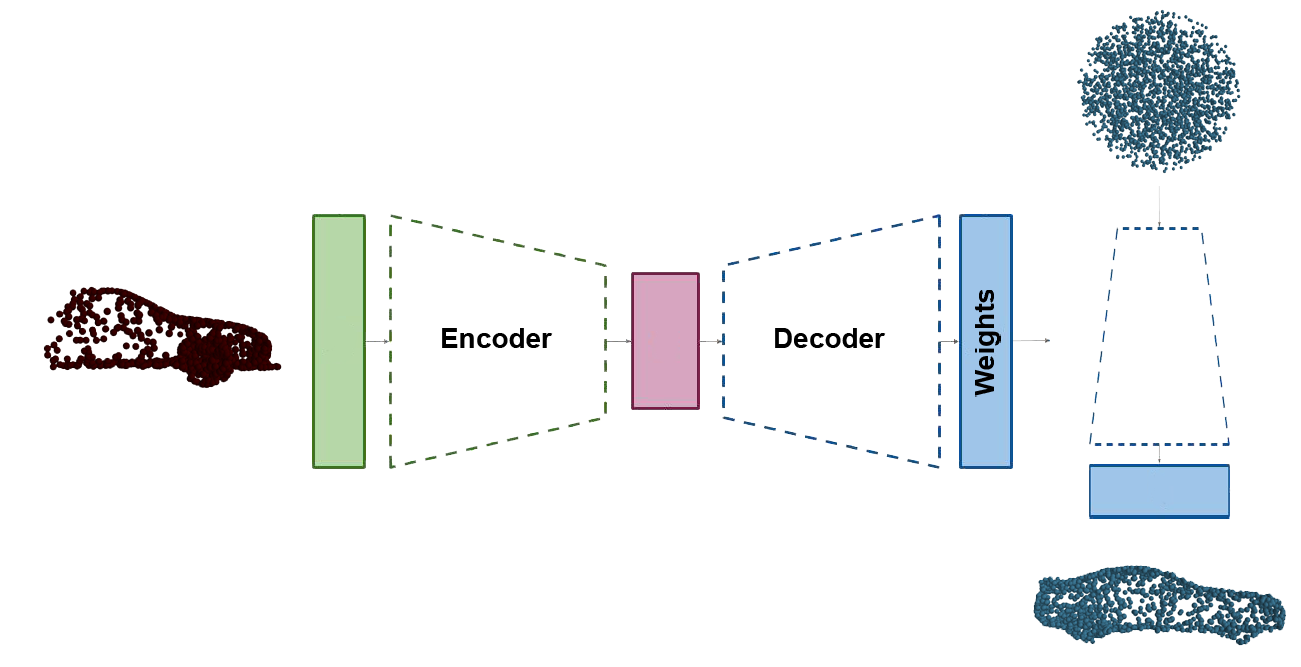} 

\end{center} 
  \vspace{-0.1cm}
  \caption{Architecture of \ours{}. This model uses a single encoder and is a dedicated for the reconstruction task.} 
\label{fig:generated_ecene_point} 
\end{figure}

\section{Appendix~B: Metrics for evaluating point cloud generation capabilities}

Let $S_g$ be the set of generated point clouds and $S_r$ be the set of reference point clouds with $|S_r| = |S_g|$. To evaluate generative models, we use the three metrics
introduced in \cite{achlioptas2018learning}:

\begin{itemize}
\item {\em Jensen-Shannon Divergence (JSD)} is computed between the marginal point distributions:
$$
JSD(P_g, P_r) = \frac{1}{2} D_{KL}(P_r||M) + \frac{1}{2} D_{KL}(P_g||M),
$$
where 
$M = \frac{1}{2} (Pr + Pg). 
$
$P_r$ and $P_g$ are marginal distributions of points in the reference and generated sets, approximated by discretizing the space into 283 voxels and assigning each point to one of them. However, it only considers the marginal point distributions but not the distribution of individual shapes. 
\item {\em Coverage (COV) }  measures the fraction of point clouds in the reference set that is matched to at least one point cloud in the generated set. For each point cloud in the generated set, its nearest neighbor in the
reference set is marked as a match:
$$
COV(S_g, S_r) = \frac{|{ \mathrm{arg min}_{ Y \in S_r} D(X, Y )|X \in S_g}|}{|Sr|},
$$
where $D$ can be either CD or EMD.

\item {\em  Minimum matching distance (MMD)} is proposed to complement coverage as a metric that measures quality. For each point cloud in the reference set, the distance to its nearest neighbor in the generated set is computed and averaged:
$$
MMD(S_g, S_r) = \frac{1}{|S_r|}
\sum_{Y \in Sr} \mathrm{min}_{X \in  S_g} D(X, Y ),
$$
where $D$ can be either CD or EMD. 
\end{itemize}

\section{Appendix~C: Implementation details}
We implement our approach using PyTorch and CUDA. All our models use the same: target network architecture - fully connected network with following layer sizes: $3, 32, 64, 128, 64, 3$; 
Adam optimizer \cite{kingma2014adam} with an initial learning rate $0.0001$, $\beta_1 = 0.9$ and $\beta_2 = 0.999$; target network input normalization so that after $100$ epochs, the target network input is sampled from a uniform unit 3D ball.

For the Completion3D benchmark we use \ours{} architecture with StepLR scheduler (step = $41$, $\gamma=0.01$) and a latent vector $z_e$, of which the length we set to $|z_e| = 128$.

With our MissingShapeNet and 3D-EPN datasets, we use \our{} architecture with latent vectors $z_e$ and $z_m$, both with length equal to 128. To generate point clouds we produce noise to replace $\mathcal{E}_m$ output from normal distribution with $\mu = 0.0$ and $\sigma$ depends on dataset:

- chair, plane, and table categories from 3D-EPN dataset use~$\sigma = 0.13,  0.1, 0.065$, respectively.

- In experiments with MissingShapeNet, we always set $\sigma$ to $0.05$.

\section{Appendix~D: More experimental results}

In this section, we present more results split between the following figures:
\begin{itemize}
    \item Fig.~\ref{fig:generated_rec_point}: multiple variants of 3D point clouds produced by our \our{} model.
    \item Fig.~\ref{fig:two_rec_point}: the reconstructions of object parts (partial completion results) with multiple variants.
    \item Fig.~\ref{fig:adaptation_scene_1}: the process of adapting generated shapes to the geometrical constraints of the scene. 
\end{itemize}

\begin{figure*}[!h] 
\begin{center} 
 \includegraphics[width=0.9\textwidth]{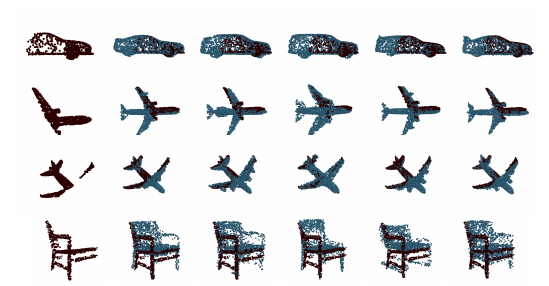}\\
\end{center} 
  \caption{Various 3D point clouds produced by our \our{} model. The first column shows an existing part of objects.
  In the next columns, we can see different versions of completions produced by our model.} 
\label{fig:generated_rec_point} 
\end{figure*}

\begin{figure*}[!h] 
\begin{center} 
 \includegraphics[width=0.9\textwidth]{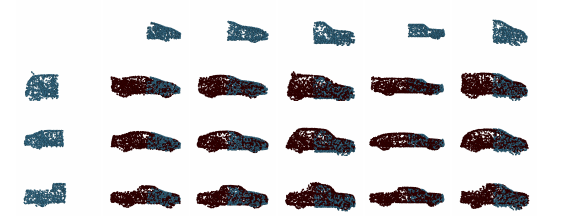}\\
\end{center} 
  \caption{Effect of reconstruction of only parts of objects. The first row presents the existing object part whereas the first column contains an input of a second encoder. The rest of the figure shows ``stitched'' reconstructions done by the \our{} model that inherit geometrical properties from inputs.} 
\label{fig:two_rec_point} 
\end{figure*}

\begin{figure*}[!ht] 
\begin{center} 
 \includegraphics[width=\textwidth]{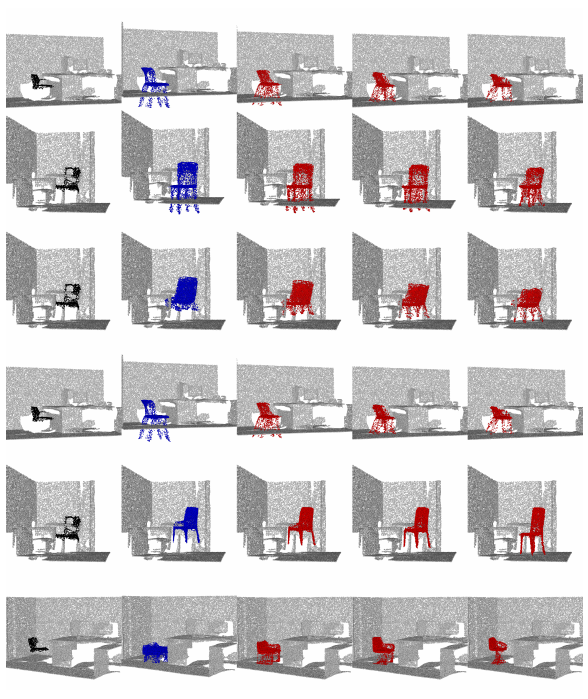}\\
\end{center} 
  \caption{Examples of chair reconstructions in scenes created by the \our{} model. The first column shows the scene with the incomplete chair (in black). The second column contains the initial reconstructions of the chair (in blue) created by the \our{} model. The next three columns present the process of adaptation.} 
\label{fig:adaptation_scene_1} 
\end{figure*}


\begin{thebibliography}{36}
\providecommand{\natexlab}[1]{#1}
\providecommand{\url}[1]{\texttt{#1}}
\expandafter\ifx\csname urlstyle\endcsname\relax
  \providecommand{\doi}[1]{doi: #1}\else
  \providecommand{\doi}{doi: \begingroup \urlstyle{rm}\Url}\fi

\bibitem[Achlioptas et~al.(2018)Achlioptas, Diamanti, Mitliagkas, and
  Guibas]{achlioptas2018learning}
Achlioptas, P., Diamanti, O., Mitliagkas, I., and Guibas, L.
\newblock Learning representations and generative models for 3d point clouds.
\newblock In \emph{International conference on machine learning}, pp.\  40--49.
  PMLR, 2018.

\bibitem[Bello et~al.(2020)Bello, Yu, and Wang]{bello2020review}
Bello, S.~A., Yu, S., and Wang, C.
\newblock Review: deep learning on {3D} point clouds.
\newblock \emph{Remote Sensing}, 12, 2020.

\bibitem[Briscoe(2011)]{briscoe2011mental}
Briscoe, R.~E.
\newblock Mental imagery and the varieties of amodal perception.
\newblock \emph{Pacific Philosophical Quarterly}, 92\penalty0 (2):\penalty0
  153--173, 2011.

\bibitem[Chang et~al.(2015)Chang, Funkhouser, Guibas, Hanrahan, Huang, Li,
  Savarese, Savva, Song, Su, Xiao, Yi, and Yu]{shapenet}
Chang, A.~X., Funkhouser, T.~A., Guibas, L.~J., Hanrahan, P., Huang, Q., Li,
  Z., Savarese, S., Savva, M., Song, S., Su, H., Xiao, J., Yi, L., and Yu, F.
\newblock Shapenet: An information-rich 3d model repository.
\newblock \emph{CoRR}, abs/1512.03012, 2015.

\bibitem[Chen et~al.(2019)Chen, Chen, and Mitra]{chen2019unpaired}
Chen, X., Chen, B., and Mitra, N.~J.
\newblock Unpaired point cloud completion on real scans using adversarial
  training.
\newblock \emph{arXiv preprint arXiv:1904.00069}, 2019.

\bibitem[Dai et~al.(2017)Dai, Ruizhongtai~Qi, and Nie{\ss}ner]{dai2017shape}
Dai, A., Ruizhongtai~Qi, C., and Nie{\ss}ner, M.
\newblock Shape completion using 3d-encoder-predictor cnns and shape synthesis.
\newblock In \emph{Proceedings of the IEEE Conference on Computer Vision and
  Pattern Recognition}, pp.\  5868--5877, 2017.

\bibitem[Goodfellow et~al.(2014)Goodfellow, Pouget-Abadie, Mirza, Xu,
  Warde-Farley, Ozair, Courville, and Bengio]{goodfellow2014generative}
Goodfellow, I., Pouget-Abadie, J., Mirza, M., Xu, B., Warde-Farley, D., Ozair,
  S., Courville, A., and Bengio, Y.
\newblock Generative adversarial nets.
\newblock In \emph{Advances in neural information processing systems}, pp.\
  2672--2680, 2014.

\bibitem[Groueix et~al.(2018)Groueix, Fisher, Kim, Russell, and
  Aubry]{groueix2018papier}
Groueix, T., Fisher, M., Kim, V.~G., Russell, B.~C., and Aubry, M.
\newblock A papier-m{\^a}ch{\'e} approach to learning 3d surface generation.
\newblock In \emph{Proceedings of the IEEE conference on computer vision and
  pattern recognition}, pp.\  216--224, 2018.

\bibitem[Ha et~al.(2016)Ha, Dai, and Le]{ha2016hypernetworks}
Ha, D., Dai, A., and Le, Q.~V.
\newblock Hypernetworks.
\newblock \emph{arXiv preprint arXiv:1609.09106}, 2016.

\bibitem[Han et~al.(2017)Han, Li, Huang, Kalogerakis, and Yu]{han2017high}
Han, X., Li, Z., Huang, H., Kalogerakis, E., and Yu, Y.
\newblock High-resolution shape completion using deep neural networks for
  global structure and local geometry inference.
\newblock In \emph{Proceedings of the IEEE international conference on computer
  vision}, pp.\  85--93, 2017.

\bibitem[Hassani \& Haley(2019)Hassani and Haley]{hassani2019unsupervised}
Hassani, K. and Haley, M.
\newblock Unsupervised multi-task feature learning on point clouds.
\newblock In \emph{Proceedings of the IEEE/CVF International Conference on
  Computer Vision}, pp.\  8160--8171, 2019.

\bibitem[Kehoe et~al.(2015)Kehoe, Patil, Abbeel, and Goldberg]{kehoe2015survey}
Kehoe, B., Patil, S., Abbeel, P., and Goldberg, K.
\newblock A survey of research on cloud robotics and automation.
\newblock \emph{IEEE Transactions on automation science and engineering},
  12\penalty0 (2):\penalty0 398--409, 2015.

\bibitem[Kingma \& Ba(2014)Kingma and Ba]{kingma2014adam}
Kingma, D.~P. and Ba, J.
\newblock Adam: A method for stochastic optimization.
\newblock \emph{arXiv preprint arXiv:1412.6980}, 2014.

\bibitem[Kingma \& Welling(2013)Kingma and Welling]{kingma2013auto}
Kingma, D.~P. and Welling, M.
\newblock Auto-encoding variational bayes.
\newblock \emph{arXiv preprint arXiv:1312.6114}, 2013.

\bibitem[Klocek et~al.(2019)Klocek, Maziarka, Wo{\l}czyk, Tabor, Nowak, and
  {\'S}mieja]{klocek2019hypernetwork}
Klocek, S., Maziarka, {\L}., Wo{\l}czyk, M., Tabor, J., Nowak, J., and
  {\'S}mieja, M.
\newblock Hypernetwork functional image representation.
\newblock In \emph{International Conference on Artificial Neural Networks},
  pp.\  496--510. Springer, 2019.

\bibitem[Lei et~al.(2019)Lei, Akhtar, and Mian]{lei2019octree}
Lei, H., Akhtar, N., and Mian, A.
\newblock Octree guided cnn with spherical kernels for 3d point clouds.
\newblock In \emph{Proceedings of the IEEE/CVF Conference on Computer Vision
  and Pattern Recognition}, pp.\  9631--9640, 2019.

\bibitem[Li et~al.(2018)Li, Bu, Sun, Wu, Di, and Chen]{li2018pointcnn}
Li, Y., Bu, R., Sun, M., Wu, W., Di, X., and Chen, B.
\newblock Pointcnn: Convolution on x-transformed points.
\newblock \emph{Advances in neural information processing systems},
  31:\penalty0 820--830, 2018.

\bibitem[Liu et~al.(2020)Liu, Sheng, Yang, Shao, and Hu]{liu2020morphing}
Liu, M., Sheng, L., Yang, S., Shao, J., and Hu, S.-M.
\newblock Morphing and sampling network for dense point cloud completion.
\newblock In \emph{Proceedings of the AAAI Conference on Artificial
  Intelligence}, volume~34, pp.\  11596--11603, 2020.

\bibitem[Mandikal \& Radhakrishnan(2019)Mandikal and
  Radhakrishnan]{mandikal2019dense}
Mandikal, P. and Radhakrishnan, V.~B.
\newblock Dense 3d point cloud reconstruction using a deep pyramid network.
\newblock In \emph{2019 IEEE Winter Conference on Applications of Computer
  Vision (WACV)}, pp.\  1052--1060. IEEE, 2019.

\bibitem[Mao et~al.(2019)Mao, Wang, and Li]{mao2019interpolated}
Mao, J., Wang, X., and Li, H.
\newblock Interpolated convolutional networks for 3d point cloud understanding.
\newblock In \emph{Proceedings of the IEEE/CVF International Conference on
  Computer Vision}, pp.\  1578--1587, 2019.

\bibitem[Mo et~al.(2019)Mo, Zhu, Chang, Yi, Tripathi, Guibas, and
  Su]{Mo_2019_CVPR}
Mo, K., Zhu, S., Chang, A.~X., Yi, L., Tripathi, S., Guibas, L.~J., and Su, H.
\newblock {PartNet}: A large-scale benchmark for fine-grained and hierarchical
  part-level {3D} object understanding.
\newblock In \emph{The IEEE Conference on Computer Vision and Pattern
  Recognition (CVPR)}, June 2019.

\bibitem[Nanay(2018)]{nanay2018importance}
Nanay, B.
\newblock The importance of amodal completion in everyday perception.
\newblock \emph{i-Perception}, 9\penalty0 (4):\penalty0 2041669518788887, 2018.

\bibitem[Qi et~al.(2017{\natexlab{a}})Qi, Su, Mo, and Guibas]{qi2017pointnet}
Qi, C.~R., Su, H., Mo, K., and Guibas, L.~J.
\newblock Pointnet: Deep learning on point sets for 3d classification and
  segmentation.
\newblock In \emph{Proceedings of the IEEE Conference on Computer Vision and
  Pattern Recognition}, pp.\  652--660, 2017{\natexlab{a}}.

\bibitem[Qi et~al.(2017{\natexlab{b}})Qi, Yi, Su, and Guibas]{qi2017pointnet++}
Qi, C.~R., Yi, L., Su, H., and Guibas, L.~J.
\newblock Pointnet++: Deep hierarchical feature learning on point sets in a
  metric space.
\newblock In \emph{Advances in neural information processing systems}, pp.\
  5099--5108, 2017{\natexlab{b}}.

\bibitem[Sheikh et~al.(2017)Sheikh, Rasul, Merentitis, and
  Bergmann]{sheikh2017stochastic}
Sheikh, A.-S., Rasul, K., Merentitis, A., and Bergmann, U.
\newblock Stochastic maximum likelihood optimization via hypernetworks.
\newblock \emph{arXiv preprint arXiv:1712.01141}, 2017.

\bibitem[Spurek et~al.(2020{\natexlab{a}})Spurek, Winczowski, Tabor, Zamorski,
  Zieba, and Trzci{\'n}ski]{spurek2020hypernetwork}
Spurek, P., Winczowski, S., Tabor, J., Zamorski, M., Zieba, M., and
  Trzci{\'n}ski, T.
\newblock Hypernetwork approach to generating point clouds.
\newblock \emph{Proceedings of the 37th International Conference on Machine
  Learning (ICML)}, 2020{\natexlab{a}}.

\bibitem[Spurek et~al.(2020{\natexlab{b}})Spurek, Zieba, Tabor, and
  Trzci{\'n}ski]{spurek2020hyperflow}
Spurek, P., Zieba, M., Tabor, J., and Trzci{\'n}ski, T.
\newblock Hyperflow: Representing 3d objects as surfaces.
\newblock \emph{arXiv preprint arXiv:2006.08710}, 2020{\natexlab{b}}.

\bibitem[Wang et~al.(2019)Wang, Chen, and Jia]{wang2019deep}
Wang, K., Chen, K., and Jia, K.
\newblock Deep cascade generation on point sets.
\newblock In \emph{IJCAI}, volume~2, pp.\ ~4, 2019.

\bibitem[Wu et~al.(2020)Wu, Chen, Zhuang, and Chen]{wu2020multimodal}
Wu, R., Chen, X., Zhuang, Y., and Chen, B.
\newblock Multimodal shape completion via conditional generative adversarial
  networks.
\newblock \emph{arXiv preprint arXiv:2003.07717}, 2020.

\bibitem[Xie et~al.(2020)Xie, Yao, Zhou, Mao, Zhang, and Sun]{xie2020grnet}
Xie, H., Yao, H., Zhou, S., Mao, J., Zhang, S., and Sun, W.
\newblock Grnet: Gridding residual network for dense point cloud completion.
\newblock \emph{arXiv preprint arXiv:2006.03761}, 2020.

\bibitem[Yang et~al.(2018{\natexlab{a}})Yang, Luo, and Urtasun]{yang2018pixor}
Yang, B., Luo, W., and Urtasun, R.
\newblock Pixor: Real-time 3d object detection from point clouds.
\newblock In \emph{Proceedings of the IEEE conference on Computer Vision and
  Pattern Recognition}, pp.\  7652--7660, 2018{\natexlab{a}}.

\bibitem[Yang et~al.(2019)Yang, Huang, Hao, Liu, Belongie, and
  Hariharan]{yang2019pointflow}
Yang, G., Huang, X., Hao, Z., Liu, M.-Y., Belongie, S., and Hariharan, B.
\newblock Pointflow: 3d point cloud generation with continuous normalizing
  flows.
\newblock In \emph{Proceedings of the IEEE International Conference on Computer
  Vision}, pp.\  4541--4550, 2019.

\bibitem[Yang et~al.(2018{\natexlab{b}})Yang, Feng, Shen, and
  Tian]{yang2018foldingnet}
Yang, Y., Feng, C., Shen, Y., and Tian, D.
\newblock Foldingnet: Point cloud auto-encoder via deep grid deformation.
\newblock In \emph{Proceedings of the IEEE Conference on Computer Vision and
  Pattern Recognition}, pp.\  206--215, 2018{\natexlab{b}}.

\bibitem[Yuan et~al.(2018)Yuan, Khot, Held, Mertz, and Hebert]{yuan2018pcn}
Yuan, W., Khot, T., Held, D., Mertz, C., and Hebert, M.
\newblock Pcn: Point completion network.
\newblock In \emph{2018 International Conference on 3D Vision (3DV)}, pp.\
  728--737. IEEE, 2018.

\bibitem[Zamorski et~al.(2018)Zamorski, Zieba, Klukowski, Nowak, Kurach,
  Stokowiec, and Trzcinski]{zamorski2018adversarial}
Zamorski, M., Zieba, M., Klukowski, P., Nowak, R., Kurach, K., Stokowiec, W.,
  and Trzcinski, T.
\newblock Adversarial autoencoders for compact representations of 3d point
  clouds.
\newblock \emph{arXiv preprint arXiv:1811.07605}, 2018.

\bibitem[Zhang et~al.(2019)Zhang, Hao, Wang, de~Silva, and Fu]{zhang2019linked}
Zhang, K., Hao, M., Wang, J., de~Silva, C.~W., and Fu, C.
\newblock Linked dynamic graph cnn: Learning on point cloud via linking
  hierarchical features.
\newblock \emph{arXiv preprint arXiv:1904.10014}, 2019.

\end{thebibliography}
\end{document}